\documentclass[10pt,twocolumn,letterpaper]{article}

\usepackage{3dv}
\usepackage{times}
\usepackage{epsfig}
\usepackage{graphicx}
\usepackage{amsmath}
\usepackage{amssymb}
\usepackage{times}
\usepackage{epsfig}
\usepackage{graphicx}
\usepackage{dblfloatfix}
\usepackage{amsmath}
\usepackage{amssymb}
\usepackage{wrapfig}
\usepackage{subcaption}
\usepackage{adjustbox}
\usepackage{comment}
\usepackage{multicol}
\usepackage{xcolor}

\usepackage[pagebackref=true,breaklinks=true,colorlinks,bookmarks=false]{hyperref}

\threedvfinalcopy % *** Uncomment this line for the final submission

\def\threedvPaperID{0079} % *** Enter the 3DV Paper ID here

\newcommand{\namelong}{Projective Urban Texturing\xspace}
\newcommand{\name}{PUT\xspace}
\newcommand{\nameA}{PUT\xspace1\xspace}
\newcommand{\nameB}{PUT\xspace2\xspace}
\newcommand{\nameC}{PUT\xspace3\xspace}
\newcommand{\nameD}{PUT\xspace4\xspace}

\ifthreedvfinal\fi
\begin{document}

%%%%%%%%% TITLE
\title{Projective Urban Texturing}

\author{Yiangos Georgiou$^{1,2}$ 
\,\,\,\,\,\,\,\, 
Melinos Averkiou$^{1,2}$ 
\,\,\,\,\,\,\,\,
Tom Kelly$^{3}$ 
\,\,\,\,\,\,\,\,
Evangelos Kalogerakis$^{4}$\\
University of Cyprus$^{1}$
\,\,\,\,\,\,\,\,
CYENS CoE$^2$
\,\,\,\,\,\,\,\,
University of Leeds$^3$
\,\,\,\,\,\,\,\,
UMass Amherst$^4$\\
%{\tt\small \{ygeorg01, maverkiou\}@ucy.ac.cy, T.W.A.Kelly@leeds.ac.uk, kalo@cs.umass.edu}
}

\twocolumn[{%
 \renewcommand\twocolumn[1][]{#1}%
 \maketitle
 \thispagestyle{empty}
 \vspace{-6mm}
 \centering
 \includegraphics[width=\textwidth]{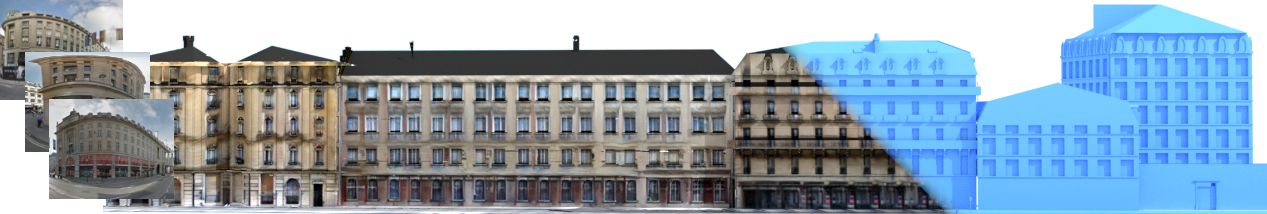}
 \vspace{-7mm}
 \captionof{figure}
  {Our method takes as input a set of panoramic images (inset, left) representative of an urban environment (e.g., a London neighborhood) along with 3D untextured polygon meshes (blue, right),  and it outputs textures for the meshes (centre) with a similar style to the input panoramic images.
  %Our method learns to transfer the textural style of panoramic image sources ( to
  % polygon meshes (blue, right), generating urban textures for them  (centre).
  \label{fig:teaser}
   }
\vspace{5mm}
}]

\maketitle
\thispagestyle{empty}

%%%%%%%%% ABSTRACT

\begin{abstract}
  This paper proposes a method for automatic generation of textures for 3D city meshes in immersive urban environments. Many recent pipelines capture or synthesize large quantities of city geometry using scanners or procedural modeling pipelines. Such geometry is intricate and realistic, however the generation of photo-realistic textures for such large scenes remains a problem. We propose to generate textures for input target 3D\ meshes driven by the textural style present in readily available datasets of panoramic photos capturing urban environments. Re-targeting such 2D  datasets to 3D  geometry is challenging because 
 the underlying shape, size, and layout of   the urban structures in the photos\ do not correspond to the ones  in the target meshes. Photos also often have objects (e.g., trees, vehicles) that may not even be present in the target  geometry. To address these issues we present a method, called \namelong (\name), which  re-targets textural style  from real-world panoramic images to unseen urban meshes. PUT relies on contrastive and adversarial training of a neural architecture designed for unpaired image-to-texture translation. The generated textures are stored in a texture atlas applied to the target 3D  mesh geometry.
 To promote texture consistency, PUT employs an iterative  procedure in which texture synthesis is conditioned on previously generated, adjacent  
textures.  %Our neural architecture generates seamless panoramic textures from geometry to project onto 3D urban models. 
  We demonstrate both quantitative and qualitative evaluation of the generated 
   textures. %-------------------------------------------------------------------------
%  ACM CCS 1998
%  (see https://www.acm.org/publications/computing-classification-system/1998)
% \begin{classification} % according to https://www.acm.org/publications/computing-classification-system/1998
% \CCScat{Computer Graphics}{I.3.3}{Picture/Image Generation}{Line and curve generation}
% \end{classification}
%-------------------------------------------------------------------------
%  ACM CCS 2012
%   (see https://www.acm.org/publications/class-2012)
%The tool at \url{http://dl.acm.org/ccs.cfm} can be used to generate
% CCS codes.
%Example:

% <concept_id>10010147.10010257.10010293.10010294</concept_id>
% <concept_desc>Computing methodologies~Neural networks</concept_desc>
% <concept_significance>300</concept_significance>
% <concept_id>10010147.10010371.10010382.10010384</concept_id>
% <concept_desc>Computing methodologies~Texturing</concept_desc>

% \ccsdesc[500]{Computing methodologies~Texturing}
% \ccsdesc[300]{Computing methodologies~Neural networks}
% \ccsdesc[500]{Methods and Applications~Neural Nets}
% \ccsdesc[300]{Imaging and Video~Texture Synthesis and Inpainting}

% \printccsdesc   

\end{abstract}

%%%%%%%%% Introduction
\vspace{-5mm}
\section{Introduction}
%\subsection{Intro}
% \subsubsection{Intro}

%We address the problem of texture generation for urban geometry. Our goal is to %generate textures for 3D building models in immersive urban enviroments  which fit %the geometry with   local or global discontinuities. We take a data-driven approach %to this problem, using unpaired training datasets of panoramic images to texture %cityscapes.

%A majority of humans live their lives in cities. 
%\melinos{this first sentence does not blend well with the previous paragraph}
Authoring realistic 3D\ urban environments is integral to city planning,  navigation, story-telling, and real-estate. Although 3D meshes of cities with  thousands or millions of polygons  can be synthesized through numerous
procedural modeling pipelines \cite{Parish:2001:PMC,Mueller:2006:PMB,kelly2011interactive,Schwarz:2015:APM,Esri:2020:CE},
they often lack texture  information.
To create vibrant and realistic virtual scenes, these meshes must be textured. Textures also provide many of the cues that we use to identify locations and purpose of buildings, including material and color. 3D\ meshes of cities can alternatively be reconstructed through mesh acquisition pipelines. However, texture generation for the acquired meshes may still be desirable especially when the captured textures have low quality and resolution
(e.g., ones from aerial photogrammetric reconstruction), or when  a different textural style is desired. Manually creating high-quality textures is time consuming -- not only must the details of the textures match the geometry, but also the style within single structures, as well as between adjacent structures and their surroundings (e.g., between buildings, street-furniture, or sidewalks). An alternative is to programmatically generate textures (e.g., using shape grammars~\cite{muller2007image}), an approach which can create repeated features (e.g., windows, pillars, or lampposts) common in urban landscapes, but is time consuming to write additional such rules. This approach is often challenged by  lack of texture variance  (e.g., due to discoloration and aging of buildings). 

We present \namelong (\name), a method to create\ street-level textures for input city meshes. Following recent texture generation work, we use a convolutional neural network with  adversarial losses to generate textures. 
However, the resolution of existing approaches is limited by available memory and they often result in numerous artifacts, as na\"ively tiling network outputs creates discontinuities (seams) between adjacent tiles (see Figure~\ref{fig:interpolation}).
%Without semantic mesh information such seams often form through prominent features.
\name iteratively textures parts of a 3D city mesh by translating panoramic street-level images of real cities using a neural network, and merges the results into a single high-resolution texture atlas. By conditioning the network outputs on previous iterations, we are able to reduce seams and promote consistent style with prior iterations. 
%Unlike temporal-consistency approaches which have 1D consistency for our goal the results must be consistent in 3D.

A central decision when developing texturing systems is the domain to learn from. Learning to create textures in 3D\ object-space (e.g., over facades)~\cite{pix2pix,kelly2018frankengan} can exploit semantic information. However, object-object interactions (e.g., between adjacent buildings or streets and buildings) are not modelled, mesh semantic information is rarely available, and mesh training data is rather sparse.
%, and fixed-resolution networks have difficulty creating consistent resolution textures over different sized objects.
In contrast, \name learns in panoramic image-space, texturing multiple objects simultaneously and modeling their interactions consistently. 

A particular advantage of panoramic image datasets is that they contain massive amounts of urban image data to train with. Advances in optics have given us low-cost commodity 360 degree panoramic cameras which capture omni-directional information quickly, while efforts such as Google's Street View have resulted in  large  panorama datasets captured in different cities \cite{anguelov2010google}. These images are typically street-level -- a single panorama captures multiple objects such opposing facades, overhanging features (e.g., roof eaves), and the street itself. 

On the other hand, there are several challenges in re-targeting  these panoramic image datasets to textures of city meshes. Since panoramic images are captured from the real world, buildings are often occluded (e.g., by trees or vehicles), and  contain camera artifacts (e.g., lens flare). Further, these\ image datasets do not contain corresponding 3D geometry.
In fact, the underlying shape, size, and layout of the urban structures in the images\ can be quite different from the ones  in the target meshes. 
Another challenge is to generate a single, consistent output texture atlas by training on an unorganized collection of images. \name addresses these challenges by employing contrastive  and adversarial learning for unpaired image-to-texture translation
along with a
texture consistency loss function. The generated textures share a similar style with the one  in the training  images e.g., training on paroramas from London neighborhoods yields London-like textures for   target meshes
(Figures \ref{fig:teaser}, \ref{fig:render_london}).

\name offers the following contributions: i) an architecture that learns to generate  textures over large urban scenes conditioned on  meshes and previous results to promote texture consistency, ii) unpaired style transfer from panoramic image datasets to texture maps in the form of texture atlases. 

%\begin{figure}[b]
%    \centering
%   \includegraphics[width=0.5\columnwidth]{imgs/pano.jpg}
% \caption{A procedurally generated scene.}
% \label{fig:multiball}
%\end{figure}
% \begin{wrapfigure}[10]{r}{0.4\columnwidth}
% \vspace*{-3ex} % 2ex is about the height of a line (twice the height of a lowercase x)
% \hspace*{-0.06\columnwidth}  % figwidth - hspace = image width
%   \includegraphics[width=0.46\columnwidth]{images/pano.jpg}
% \end{wrapfigure}

%%%%%%%%% Related Work
%\vspace{-1mm}
\section{Related Work}
\label{related}

\begin{comment}
\paragraph{Urban 3D Mesh Synthesis.}
Virtual urban environments have become ubiquitous for city planning, video game environments, and mapping. Following this trend there are increasing numbers of 3D city model sources - both for reconstruction~\cite{musialski2013survey,baillard1999automatic,zhu2018large} of existing environments and \emph{synthesis} of designed, planned, or fictional cityscapes. Synthesis techniques for urban geometry may be manual or procedural. In a manual pipeline, designers build meshes using specialised software~\cite{lipp2014pushpull++} or guided by images~\cite{nan2010smartboxes}. Such manual approaches can be slow for massive urban areas. A faster approach is procedural modeling with approaches such as Split Grammars~\cite{wonka2003instant, Mueller:2006:PMB, demir2014proceduralization} -- these models can be driven by various goals such as optimisation~\cite{Schwarz:2015:APM}, inverse procedural modeling~\cite{Wu:2014:IPM, Martinovic:2013:BGL}, or directly from real world data~\cite{kelly2018frankengan}. Procedural languages have focused on the generation of geometry -- texturing has been an afterthought. %Synthesized urban meshes have a wide variety of characteristics due to the variety of goals, approaches, data (resolution, accuracy, and range), and techniques (aerial, street-side, procedural, vehicle-mounted).
\end{comment}

%\subsection{Texture Generation \& Blending}
\name is related to methods for 2D texture synthesis and  texture generation  for 3D objects as briefly discussed below. 

%Textures are an essential component of modeling urban appearances. They add visual %information about geometric and material appearance~\cite{gruber1995geometric}.
 %and approximate high frequency geometric information
%

\textbf{Texture synthesis.} 
%is the generation of texture images from exemplars or lower dimension representations. 
Traditional learning-based methods generate textures with local and stationary properties~\cite{efros1999texture,WeiTree,LiangPatch, efros2001image, li2015approximate}. Later work creates larger, more varied textures quickly~\cite{darabi2012image}, some of which are domain specific~\cite{dai2013example}. Procedural languages can create homogeneous textures for urban domains~\cite{muller2007image}; however, they lack realism since they do not model the heterogeneous appearance of surfaces due to weathering and damage. Such phenomena can be created using simulation or grammars~\cite{chang2003synthesis, chen2005visual, PE:VMV:VMV12:063-070,lu2007context}, yet these approaches require significant hand-engineering. 
With the advent of convnets (CNNs),  data-driven texture synthesis has been achieved by optimization in feature space~\cite{gatys2015texture} or with GANs~\cite{ZhouTextureSynthesis,fruhstuck2019tilegan}. Image-to-image translation with  U-Nets~\cite{ronneberger2015u,pix2pix} has been demonstrated for style transfer~\cite{gatys2016image}; these have been applied to various domains~\cite{guerin2017interactive,kelly2018frankengan,zhang2020synthesis}. Processing panoramic images, such as street-view images, with CNNs has been explored for depth prediction~\cite{sharma2019unsupervised} and image completion~\cite{song2018im2pano3d}. Approaches for unpaired image translation have also been developed using cycle consistency losses~\cite{cycleGAN}, which learn to translate from one image domain to another, and back again. These techniques has been extended to multi-domain~\cite{choi2018stargan,huang2018multimodal} and domain-specific problems~\cite{guo2019gan, wang2018video}. A more recent approach uses patch-wise contrastive learning~\cite{CUT}. We adopt this approach as our backbone and incorporate it within our  iterative pipeline that conditions  synthesis on input 3D meshes and previous texture results, along with a novel  texture consistency loss.

\textbf{Texturing 3D objects}. Early approaches directly project photos onto reconstructed geometry~\cite{niem1995mapping, bohm2004multi,bornik2001high}. This produces photorealistic results, but requires exact photo locations with consistent geometry (i.e., noise-free location data~\cite{klingner2013street}) and unobstructed views (i.e., no occlusions by trees or vehicles~\cite{FanCompletion}).  3D CNNs have been proposed to generate voxel colours~\cite{oechsle2019texture} as well as geometry~\cite{wu2016learning, wen2019pixel2mesh++}, however such have high memory requirements and there are sparse training 3D\ object datasets with realistic textures. 
Attempts to overcome these limitations are ongoing, using a variety of approaches~\cite{kato2018neural,hedman2018deep,henzler2020neuraltexture,meshry2019neural,Raj_2019_CVPR_Workshops,henderson2020leveraging}; results are generally low resolution or only in screen-space.
GANs have been previously proposed to align and optimize a set of images for a scanned object~\cite{huang2020adversarial} producing high quality results; however, they require images of a specific object.
In our case, we leverage large 2D panoramic image datasets for generating textures for urban models without paired image input.
A key difference over prior work is that our method learns textures by combining contrastive learning-based image synthesis~\cite{CUT} with a novel blending approach to wrap the texture over the geometry.

\section{Method}

\begin{figure*}[t!]
    \centering
    \includegraphics[width=2\columnwidth]{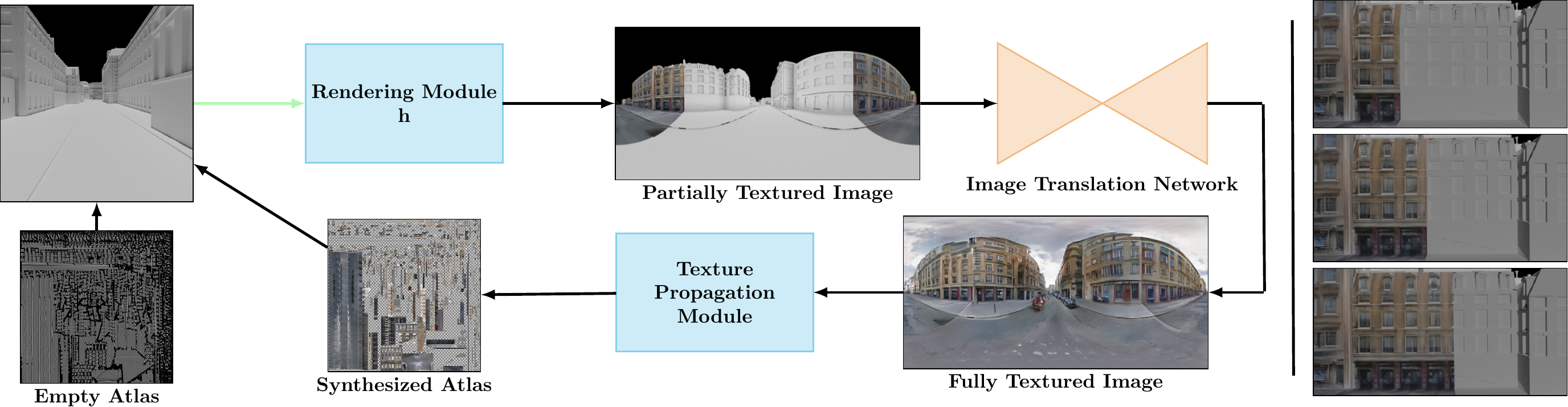}
    \vspace{-2mm}
    \caption{At each iteration our system creates a \textit{partially textured image} by rendering the 3D geometry(top left) using the \textit{synthesized atlas} containing texture from previous iterations. Mesh parts that have not been textured so far are rendered in grayscale.
The \textit{texture propagation module} uses the generated image to update the texture atlas used in the next iteration. A visualization of our iterative texturing synthesis output is displayed on a characteristic building facade on the right.} 
    \label{fig:Texturing_pipeline}
    \vspace{-5mm}
\end{figure*}

\paragraph{Overview.}
%\label{sec:overview}

Given untextured 3D geometry representing an urban environment as input, our pipeline (Figure \ref{fig:Texturing_pipeline}) synthesizes a texture for it.  We assume that the surface of the input 3D geometry is unwrapped (parameterized), such that UV coordinates assign 3D\ surface points to 2D\ locations within a texture atlas (a 2D image map; Figure \ref{fig:Texturing_pipeline}, bottom left). The input urban geometry may come from a variety of sources, such as procedural modeling (Figure \ref{fig:teaser}, right) or photogrammetric reconstruction (Figure \ref{fig:google_text}, right). % cityengine and google earth images.
%The input urban environment is generated with procedural modeling techniques, and contains buildings, roads, trees, and cars. 

Our texture synthesis system is iterative, working sequentially with selected viewpoints along street center-lines: to begin each iteration, our \emph{rendering module} creates a panoramic image representing the input 3D geometry from a viewpoint, including any previously generated texture (Figure \ref{fig:Texturing_pipeline}). The rendered images are \emph{partially textured}, since some parts of the 3D urban models may have associated texture synthesized from previous iterations, while others may not contain any texture at all. Untextured parts are rendered in grayscale, and textured parts in colour. 

As each iteration continues, the  partially textured  image is translated into a fully textured RGB\ image through a neural network (Figure \ref{fig:Texturing_pipeline}, \emph{image translation} network). Our experiments found that passing the  partially textured images through the network offered better performance, compared to passing completely untextured (grayscale) images, or using separate images, or channels, for the textured and untextured parts. The network is trained on a set of panoramic RGB images taken from streets of a particular city (e.g., London or New York) and a set of rendered images of urban geometry meshes. The two sets are unpaired; we do not assume that the urban geometry meshes have corresponding, or reference, textures. Such a level of supervision would require enormous manual effort to create training textures for each 3D\ urban model. Our system is instead trained to automatically translate the domain, or geographic ``style'', of the panoramic images of real-world buildings into textures that match the geometry ``structure'' of the input 3D urban models, and any prior textured parts. 

\begin{figure*}[t!]
    % \centering
    % \begin{subfigure}[b]{\columnwidth}
        % \centering
        % \includegraphics[width=1\columnwidth]{imgs/render_london_1_res.jpg}
    % \end{subfigure}
    % \hfill
    \begin{subfigure}[b]{\columnwidth}
        \centering
        \includegraphics[width=1\columnwidth]{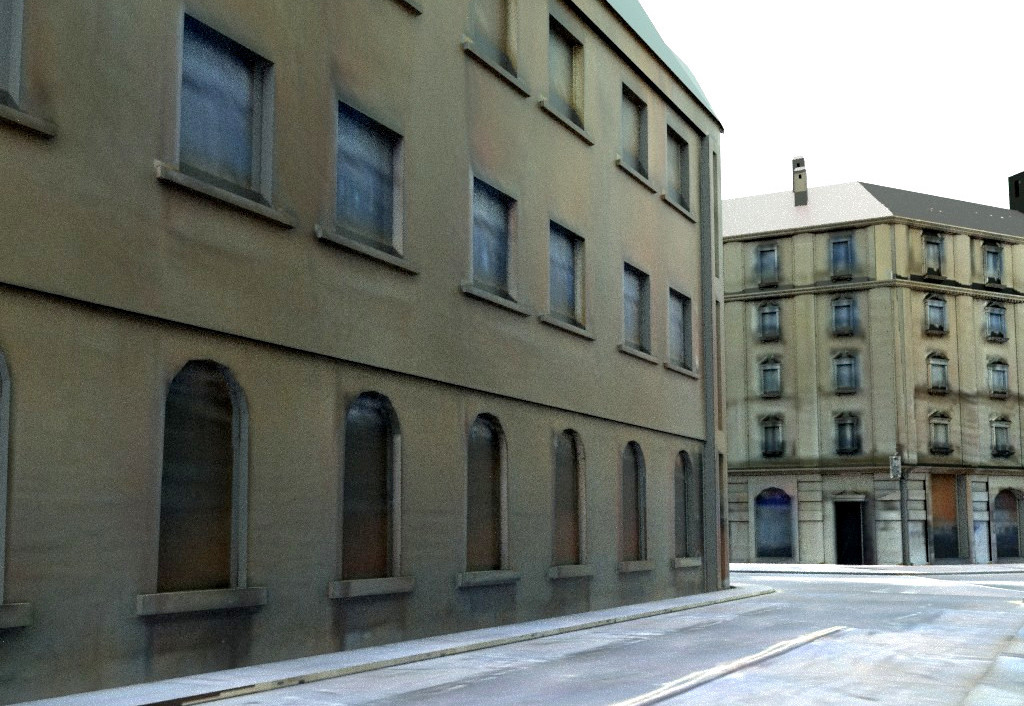}
    \end{subfigure}
    \hfill
    \begin{subfigure}[b]{\columnwidth}
        \centering
        \includegraphics[width=1\columnwidth]{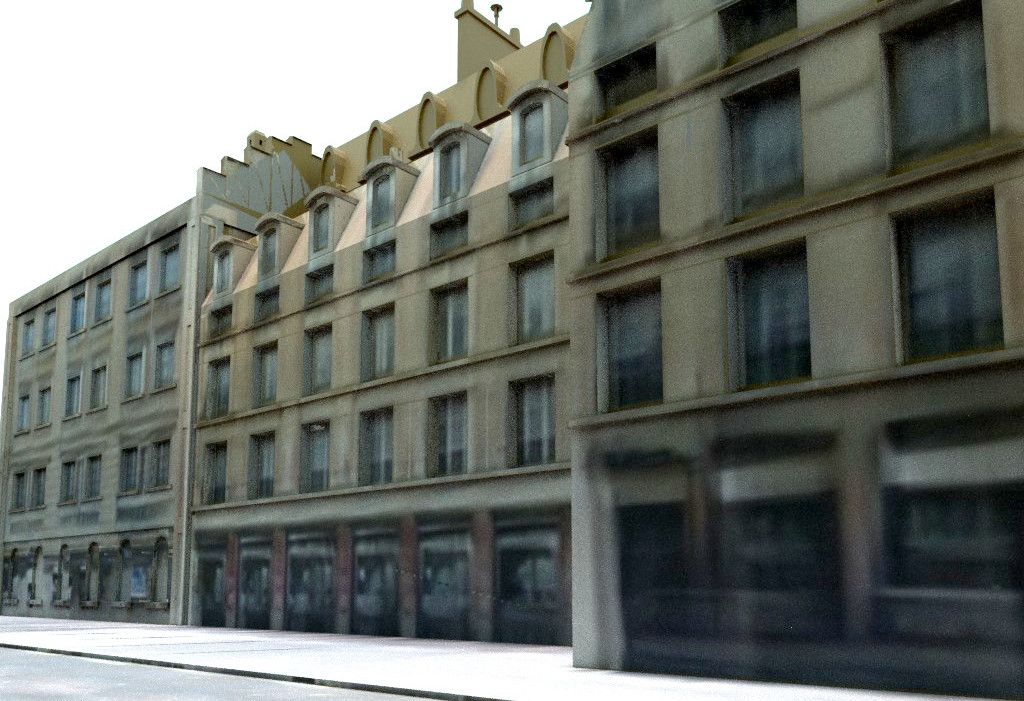}
    \end{subfigure}
    \begin{subfigure}[b]{\columnwidth}
        \centering
        \includegraphics[width=1\columnwidth]{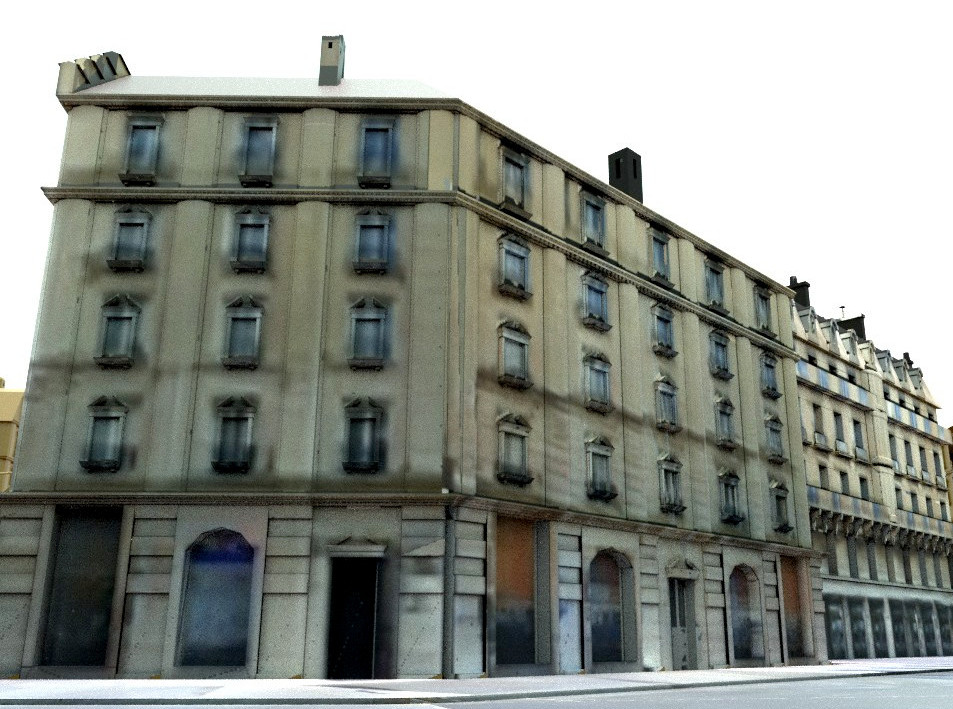}
    \end{subfigure}
    \hfill
    \begin{subfigure}[b]{\columnwidth}
        \centering
        \includegraphics[width=1\columnwidth]{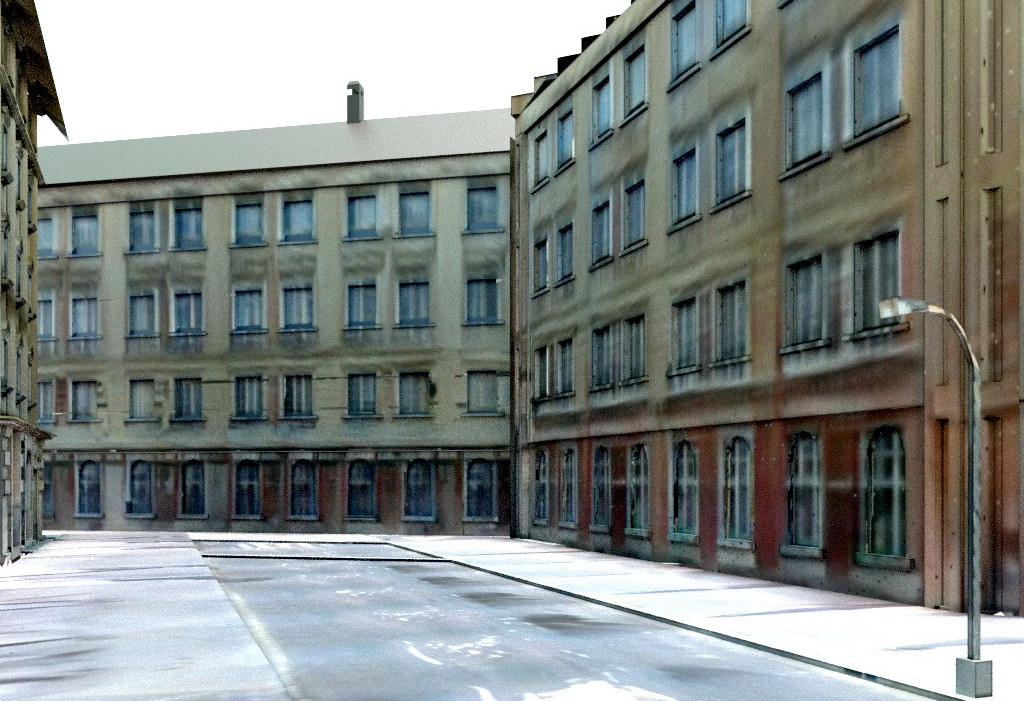}
    \end{subfigure}
    % \hfill
    % \begin{subfigure}[b]{\columnwidth}
        % \centering
        % \includegraphics[width=1\columnwidth]{imgs/render_london_7_res.jpg}
    % \end{subfigure}
    % \begin{subfigure}[b]{\columnwidth}
    %     \centering
    %     \includegraphics[width=1\columnwidth]{imgs/comparison_figure_London.png}
    % \end{subfigure}
    % \hfill
    % \begin{subfigure}[b]{\columnwidth}
    %     \centering
    %     \includegraphics[width=1\columnwidth]{imgs/comparison_figure_NY.png}
    % \end{subfigure}
    \vspace{-3mm}
    \caption{Renderings of our test scene textured by \name trained on London panoramas. We show street-level renderings.}
    \label{fig:render_london}
    \vspace{-3mm}
\end{figure*}

At the end of the iteration, our \emph{texture propagation module}
updates the texture atlas using the output fully textured image from the network (Figure \ref{fig:Texturing_pipeline}, \emph{synthesized atlas}). After the final iteration, all viewpoints have been processed, the texture atlas is complete, and is used to render the urban environment, as shown in Figure \ref{fig:render_london}.

%\yiangos{Maybe we can remove this}In the following Section \ref{sec:pipeline}, we describe the main modules of our pipeline: the rendering module for creating the partially textured images, the image translation network that generates the fully texture imaged, and the texture propagation module that updates the texture atlas. In Section \ref{sec:training}, we discuss our training datasets, and the procedure to train our neural network. 
%\section{Pipeline}
%\label{sec:pipeline}
%Our pipeline synthesizes texture for the input 3D geometry iteratively. In each iteration, we create renderings capturing the input city geometry, including any partial textures from previous passes, then the partially textured renderings are translated into complete RGB images by a neural network.

\vspace{-5mm}
\paragraph*{Viewpoint selection.} In a pre-processing step, we create paths along which we place viewpoints to sequentially capture the input geometry of the city blocks. The paths are formed by following the centerlines of all streets contained in the urban 3D scene (see Section~\ref{sec:training}). The paths trace the street from the start to the end, ``sweeping'' the buildings in each block on both sides of the road. Each viewpoint is placed at 5m horizontal intervals along the street centerlines with a vertical height of $2.5$m. The horizontal intervals are empirically selected such that each rendered image has an overlap (about $25\%$)  with the image rendered from the previous and next viewpoint. The vertical height is motivated by the fact that real-world car mounted cameras are placed at a height close to $2.5$m, reducing 
the ``domain gap'' between our renderings and real-world panoramas taken from cars. The viewpoint (camera) up and forward axes are set such that they are aligned with the world upright axis and street direction respectively. As a result, the cameras ``look'' towards the buildings along both sides of the road.

\vspace{-5mm}
\paragraph*{Rendering.}
Given each viewpoint, each rendering is produced through equirectangular (panoramic) projection. As a result, the domain gap with real-world panoramas is decreased with respect to the projection type. The geometry is rendered with global illumination using Blender \cite{blender} at a $512$x$256$ resolution. Any textured parts from previous passes incorporate RGB color information from the texture atlas, while for untextured parts, a white material is used, resulting in a grayscale appearance.
Background (e.g., sky) is rendered as black. An example of the resulting ``partially textured'' image is shown in Figure \ref{fig:Texturing_pipeline}.  
%We also experimented with a different technique, where we render the whole input geometry as grayscale into one image, and render only the textured parts into a separate RGB image (untextured parts were rendered as black). \yiangos{remove the rest?}Using these images as input to the network resulted in worse performance. Feeding one single image merging both untextured and textured parts improved the consistency of the generated textures across different viewpoints.

\vspace{-5mm}
\paragraph*{Neural network.} The input to our image translation network is a partially textured rendering ($3$x$512$x$256$) at each iteration. The neural network uses the architecture from Johnson et al.~\cite{Johnson}, also used in Contrastive Unpaired Translation (CUT)~\cite{CUT}. 
The network contains three convolution layers, 9 residual blocks, two fractionally-strided convolution layers with stride $1/2$, and a final convolution layer that maps the features to a RGB image ($3$x$512$x$256$). 
Since our input is a panoramic image, we use equirectangular convolution for the first convolutional layer of the network~\cite{equi_conv}. Equirectangular convolution is better suited for processing panoramic images and offers better performance in our experiments.
As discussed in Section~\ref{sec:training}, we follow a training procedure and loss inspired by \cite{CUT}, yet with important differences to handle partially textured images and ensure consistency in the texture generated from different viewpoints.

\begin{figure}[!ht]
    \centering
    \vspace{-2mm}
    \begin{subfigure}[b]{0.49\columnwidth}
        \centering
        \includegraphics[width=1\columnwidth]{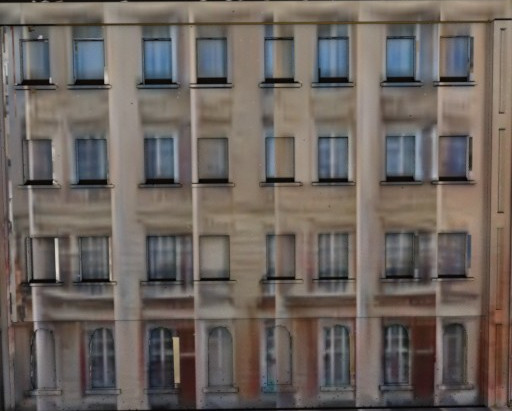}
    \end{subfigure}\hfill
    \begin{subfigure}[b]{0.49\columnwidth}
        \centering
         \includegraphics[width=1\columnwidth]{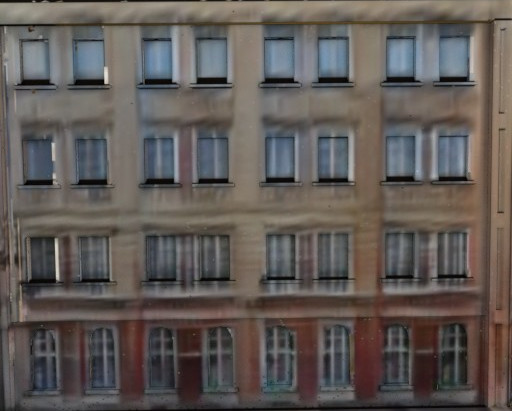}
    \end{subfigure}
    \vspace{-2mm}
    \caption{Texture w/o blending (left), w/ blending (right)}
    \label{fig:interpolation}
    \vspace{-2mm}
\end{figure}

\vspace{-5mm}
\paragraph*{Texture propagation.}\label{subsec:textprob}
% \setlength{\columnsep}{10pt}
% \begin{wrapfigure}{R}{0.5\linewidth}
%  \vspace{-2mm} 
%   \includegraphics[width=1\linewidth]{imgs/blending.pdf}
%   \vspace{-6mm}   
%   \caption{Distances of a point to projection centers of two cameras used for texture propagation.}
% \vspace{-2mm}     
%   \label{fig:blending}
% \end{wrapfigure}
At each iteration, we use the generated image from the above network to update the texture atlas. 
For each texel $t$ with center coordinates $(u_t,v_t)$ in the atlas, we find the corresponding 3D point $\mathbf{p}_{t}$ on the input geometry. Then for this 3D point, we find its corresponding pixel location in the generated image at the current iteration  $i$:
$[x_{t,i}, \, y_{t,i}]=\Pi_i(\mathbf{p_t})$, where $\Pi_i$ is the equirectangular projection function used during the frame rendering of the current iteration. The color of the texel is  transferred with the  mapping: $color[u_t,v_t]~=~\mathbf{R}_i[ x_{t,i}, \, y_{t,i} ]$, where $\mathbf{R}_i$ is the generated image from the network at the current iteration. However, this strategy can\ create artifacts since it will overwrite the color of texels that were  updated from previous iterations, resulting in sharp discontinuities in the generated texture (Figure \ref{fig:interpolation}, left).

 Instead, we follow a pooling strategy, where colors for each texel are aggregated, or blended, from different iterations. Specifically, at each iteration, the color of each texel is determined as a weighted average of pixel colors originating from  images generated from  previous iterations:
\begin{equation} \label{eq:interp}
\vspace{-2.5mm}
color[u_t,v_t] = \sum\limits_{i \in V_t} w_{i,t} \mathbf{R}_i[ x_{t,i}, \, y_{t,i}]
% \vspace{-1mm}
\end{equation}
where $V_t$ is the set of iterations where the texel's  corresponding 3D\ point $\mathbf{p}_{t}$ was accessible (visible) from the cameras associated with these iterations. The blending weights $w_{i,t}$ are determined by how close $\mathbf{p}_t$ was to the center of the projection at each iteration. The closer to the centre the point $\mathbf{p}_t$ was, the higher the weight was set for the color of its corresponding pixel at the generated image.
Specifically, if $d_{i,t}$ is the distance of the point $\mathbf{p}_t$ to the projection centerline for the camera at iteration $i$, the weight of its corresponding pixel color from the generated image at this iteration was determined as $w_{i,t}=1-d_{i,t}/\sum_{i'}d_{i',t}$, where $i'$ are  iteration indices where the point was visible. 
The weights were further clamped to $[0.3, 0.7]$, then were re-normalized between [0, 1] to eliminate contributions from cameras located too far away from the point.
%At the final iteration, the texture atlas is the result of aggregating over all iterations.  
% \kalo{(a) i wrote it in the general setting of any N of cameras (more than 2). I\ know our implementation is specific to 2, but it's better to have it as general as possible e.g., if one went with much denser viewpoints
% (b) The original formulas had too much detail. Instead of writing a big branch, weight clamping should do the equivalent job... you may change the value 0.1 depending on the threshold you set. (c) The old algorithm 1 is not needed  / too verbose. The image was created in illustrator - feel free to change the background i.e., different building - it's easy to do it, you can send it to me if you want. Adjust the wrapfigure position later. }

%\vspace{-2.5mm}
\section{Training}
\label{sec:training}

% \begin{figure}[t]
%     \centering
%     \includegraphics[width=1\columnwidth]{imgs/multiball.jpg}
%     \caption{Procedurally created urban geometry used in our method.}
%     \label{fig:multiball}
% \end{figure}

\begin{figure*}[t!]
    \centering
    \includegraphics[width=\textwidth]{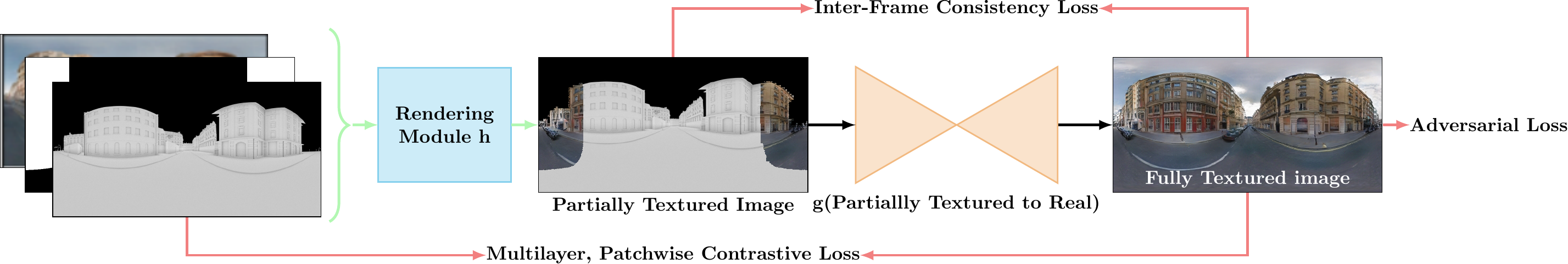}
    
    \def\svgwidth{\linewidth}
    \vspace{-2mm}
    % \caption{\name architecture used during training. The green arrow shows the mapping $h$ from a grayscale rendering of a training city block to the partially textured image. This mapping is performed in our rendering module that integrates any previous texture information with the help of a binary mask indicating prior textured regions.  The image translation network $g$ maps the partially textured image to a fully textured one, mimicking real-world statistics with the help of an adversarial loss. The partially-textured image is compared to the fully textured image in the inter-frame consistency loss to discourage discontinuities in the output texture.}
    \caption{Training procedure: the green arrow shows the mapping $h$ from a grayscale rendering of a training city block to the partially textured image. This mapping is performed in our rendering module that integrates previous texture information with the help of a binary mask indicating prior textured regions.  The image translation network $g$ maps the partially textured image to a fully textured one, mimicking real-world statistics with the help of an adversarial loss. The partially-textured image is compared to the fully textured image in the inter-frame consistency loss to discourage seams in the output texture.}
    \label{fig:PUT-Net_F2R}
    \vspace{-5mm}
\end{figure*}

To train the image translation neural network of our pipeline, we require a set of training examples from both domains: rendered panorama images from the (untextured) 3D geometry, and real-world panoramic photographs. The two sets are unpaired, i.e., the real-world images have no geometric or semantic correspondences with rendered panoramas.  We discuss the datasets used for these domains in the following paragraphs, then explain our training procedure. 

%\subsection{Datasets}\label{subsec:train_data}
\vspace{-5mm}
\paragraph*{Real-world panorama dataset.} For real-world panoramas, we use the  dataset from \cite{mirowski2019streetlearn}, with $29,414$ images from \emph{London} and $6,300$ images from  \emph{New York}; these photos are taken from city streets with vehicle mounted cameras. 

\vspace{-5mm}
\paragraph*{Rendered panoramas from 3D geometry.}
We demonstrate our system on two classes of 3D urban meshes - procedural and photogrammetrically reconstructed meshes.
%Note that the technique is applicable to any urban mesh. %Our 3D meshes were created using a procedural modeling pipeline.
%\yiangos{Elaborate or mention the procedural modeling pipeline here?}
%\twak{no we say less in case Purdue reads it again}
%(Figure \ref{fig:multiball}). 
%To ensure correct correspondences between domains it was important that the distributions of features in the images was similar to our real-world ground truth.  was applied to procedurally generate streets~\cite{Parish:2001:PMC} and parcels~\cite{vanegas2012procedural} 
Procedural meshes are generated using the CGA language~\cite{Mueller:2006:PMB}; each generated scene has unique street graph, block-subdivision, and building parameters (e.g., style, feature size, height etc.). We found that the inclusion of details such as pedestrians, trees, and cars helped the multi-layer patch-wise contrastive loss used in our network to identify meaningful correspondences between the real and rendered panorama domains.
We also export the street center-line data from the procedural models (no other ``semantic'' information or labels are used).
We generated $10$ scenes with $1600$ viewpoints sampled from $18$ random paths on streets of each scene. This resulted in $16K$ viewpoints in total, from which we rendered grayscale images of the input geometry using equirectangular projection.
We use $9$ scenes ($14.4K$ grayscale, rendered panoramic images) for training, and keep one scene for testing ($1.6K$ images). We note that the streets, building arrangements, and their geometry are unique for each street and scene, thus, the test synthetic renderings were different from the training ones.

The photogrammetric mesh dataset is from Google Earth. 
%While visually rich in the browser, 
The 3D meshes are coarse and noisy (Fig. \ref{fig:google_text}, right, blue mesh). They contain holes, vehicles, and trees. This contrasts the precise, high-detail, data in procedural meshes. Street centerlines are collected from OpenStreetMap. We use $2$ scenes and $1,411$ rendered images. One is used for training and the other for testing. 

\vspace{-5mm}
\paragraph{Training Procedure.}%\label{subsec:training}
As mentioned above, since there are no target reference texture images for the input 3D geometry, we resort to a training procedure that uses the unpaired real-world RGB panoramas and our partially textured panorama renderings. Contrastive Unpaired Translation (CUT) \cite{CUT} is a fast and flexible method to train image translation networks between unpaired domains. One potential strategy is to use CUT's losses and training as-is in our setting. However, this strategy disregards the need for consistency between  generated textures at different iterations of our pipeline.
%i.e., any 3D geometry accessible should be textured as consistently as possible  across different viewpoints. 
Further, our image translation network processes partially textured images, which are different from both the domain of real-world panoramas and the domain of grayscale rendered images. 
Next, we explain how our training procedure promotes consistency and handles the different kinds of inputs required for our network.
\vspace{-6mm}
\paragraph*{Multi-layer Patch-wise Contrastive Loss.}
The goal of our training procedure is to  learn the parameters of our image translation network $g: P\to R$ mapping the domain $P$ of partially textured images to the domain $R$ of fully textured ones mimicking real-world statistics. To associate input and output structure we use a multi-layer patch-wise contrastive loss between input and output. The association between images is achieved by the comparison of the stack of features for selected layers in the encoder network $f$. 
% Two layer MLP network $m$ is used to create fixed sized stack features given selected layers' feature map outputs.

Specifically, in our setting, given an untextured (grayscale) rendering $S$ from our set of untextured renderings $\mathbf{S}$, 
we first convert it to a partially textured one by integrating previously textured parts indicated from a binary mask $M$ ($1$ for previously textured parts, $0$ for untextured ones). This conversion $h: S \to P $ is a fixed, parameter-free mapping implemented in our rendering module.
Given a rendered image $S$ and the generated image $g(h(S))$, we generate their stack of features \mbox{$z = f(S)$} and \mbox{$\hat{z} = {f(g(h(S)))}$}. Pairs of patches from the rendered and generated images at the same location $n$ are treated as positive, while pairs from different locations are treated as negative. A temperature-scaled cross-entropy loss $H$ (with temperature $\tau=0.07$) computes the probability of the positive example being selected over negatives \cite{CUT}. The loss is summed over our dataset and samples:
%\vspace{-1.5mm}
%\begin{align*}\label{eq:cross-entro}
    %ce(u, u^+, u^-) &= -log[\frac{exp(u \cdot u^+/r)}{exp(u \cdot %u^+/r)+\sum_{n=1}^N exp(u \cdot u^-/r)}] \\
    %\mathcal{L}_{contrastive_P} &=\,\, \mathbb{E}_{ \mathbf{S} \sim p_{data}(S)}\sum_{l=1}^L\sum_{n=1}^N(ce(\hat{z_l^s}, z_l^s, z_l^{S/s})
%\end{align*}
\begin{equation}
    \mathcal{L}_{contr_S} =\,\, \mathbb{E}_{ 
    S \sim \mathbf{S}}
    \sum_{l=1}^L\sum_{n=1}^{N_l}
    H(\hat{z_l^n}, z_l^n, z_l^{N_l/n})
\end{equation}
where $L$ is the number of selected layers and $N_l$ is the index set of spatial locations in feature maps at each layer. As in CUT~\cite{CUT}, to avoid unnecessary generator changes, a cross-entropy loss $\mathcal{L}_{contr_R}$ is also applied on patches sampled from the real images domain $\mathcal{R}$.
%\begin{align*}
%    \mathcal{L}_{contrastive_R} &=\,\, \mathbb{E}_{ \mathbf{R} \sim %p_{data}(R)}\sum_{l=1}^L\sum_{n=1}^{N_l}
%    ce(\hat{r_l^s}, r_l^s, r_l^{S/s})
%\end{align*}
%where $\hat{r} = f(g(R))$ and $r = f(R)$. 

% where $\mathbf{R}$ and $\mathbf{S}$ are training examples sampled from the domain of real-world panoramas, and grayscale renderings respectively.
% We note that the real-world panoramas are not associated with any mesh, thus, the generated grayscale images from the network $f$ do not have any predetermined masks. To implement the cycle starting from the real-world panoramas, we generate synthetic 2D binary masks using equirectangular projections of rectangles, simulating actual masks. For values of the synthetic mask equal to zero we use the generated grayscale images from the network $f$, while for value of one we use the texture from the corresponding real-world images for this cycle.

\vspace{-5mm}
\paragraph*{Adversarial losses.}
Apart from the contrastive losses, we use an adversarial loss to ensure that the fully textured images generated from the image translation network $g$ share similar statistics with the real-world panoramas:
\begin{align*}
    \mathcal{L}_{gan} &=
    \mathbb{E}_{ R \sim \mathbf{R}} [log(d_r( R ))] \\
    &+ \mathbb{E}_{ S \sim \mathbf{S}} [1 - log(d_r(g(h( S )))]
%  \label{eqn}
    % \stepcounter
\end{align*}
% \begin{align*}
%     \mathcal{L}_{gan,s} &=
%     \mathbb{E}_{ \mathbf{S} \sim p_{data}(S)} [log(d_s(\mathbf{S} ))] \\
%  &+ \mathbb{E}_{ \mathbf{R} \sim p_{data}(R)} [1 - log(d_s(f( \mathbf{R}))]
% \end{align*}
where $\mathbf{R}$ is the training set of real-world panoramas,  $d_r$ is a discriminator network following the architecture of CUT applied to real images domain.

\vspace{-5mm}
\paragraph*{Inter-frame consistency loss.}
We introduce an inter-frame consistency loss to promote consistency in the generated textured images from our image translation network $g$ across different iterations (``frames'') of our pipeline. Given a fully textured image $g(h(S))$ generated from the image translation network during training, and a partially textured image $h(S)$ containing texture from prior iterations, we compare the textured regions in $h(S)$ with the corresponding generated regions. The comparison is performed with the help of the binary mask $M$ containing 1s for the partially textured regions of $h(S)$ and 0s otherwise. Using the above-mentioned mappings, the loss is expressed as follows:
\begin{equation}
    \mathcal{L}_{cons} = \mathbb{E}_{S \sim \mathbf{S}}[\parallel(g(h(S))) \odot M -  h( S ) \odot M \parallel_1]
    \label{eq:cons}
\end{equation}
where $\odot$ is the Hadamard product.
We discuss the effect of this loss in our experiments and ablation study.

\vspace{-4mm}

\paragraph*{Full Objective.}
The full objective is a weighted combination of the above losses:
\begin{equation}
    \mathcal{L} = \lambda_1 \mathcal{L}_{gan} + \lambda_2 \mathcal{L}_{cons} + \lambda_3 \mathcal{L}_{contr_S} + \lambda_4 \mathcal{L}_{contr_R}
\end{equation}
where $\lambda_1, \lambda_2, \lambda_3, \lambda_4$ are hyper-parameters set to $1.0,10.0,$ $0.5,0.5$ respectively.

\vspace{-5mm}
\paragraph*{Implementation details.} We use the Adam optimizer with learning rate $2\cdot10^{-4}$ to train the above architecture. We train the model for $200$ epochs. During the first $150$ epochs the learning rate is fixed, then it gradually decays for the last $50$ epochs. 
All our code and datasets, including the rule-set and scripts for the  procedural generation of the training and test scenes are available on our project page 
\url{https://ygeorg01.github.io/PUT/}
.

\section{Evaluation}
\label{sec:eval}

We now evaluate our pipeline on test models generated with the process described in Section \ref{sec:training}.
As discussed earlier, no identical city blocks exist in our train and test meshes. 
We trained our translation network separately on the images datasets of \emph{London} and \emph{New York}. 
%The evaluation was performed on a single procedural scene; this was excluded from the training data. 

%the $10_{th}$

% Change Accordingly the final results
% We present qualitative as well as quantitative results.
% Qualitative results include both intermediate outputs (generated panoramas)  of our image translation network as well as perspective renderings of our test meshes at street level.
% They aim to evaluate our method's ability to generate consistent and plausible textures for an urban mesh, as well as its ability to transfer the distinctive ``style'' of a real city to a virtual one.
% Quantitative results include an ablation study of the effects of various parts of our pipeline using the Fr\'echet Inception Distance (FID)\cite{fid_paper}, as well as a user study that aims to perceptually evaluate whether our output textures are more plausible than those generated by the original CycleGAN method \cite{cycleGAN}.

\vspace{-5.5mm}
\paragraph{Qualitative Evaluation.}
First, we evaluate the quality of our  output textures by placing perspective cameras at the street level for each of our test meshes and rendering the meshes with texture atlases created by our method.
Results can be seen in Figure \ref{fig:render_london}, for the network trained on London data. 
%A comparison of the styles generated by London and New York datasets are presented in Figure \ref{fig:style_lo_ny}.
Our method transfers the appearance of large structures such as multi-storey walls, streets, or sidewalks from real panoramas and often aligns texture details with geometric ones i.e., texture edges of windows and doors with geometry edges. A side-effect of our method is that shadows are sometimes translated to streets and sidewalks (Figure \ref{fig:render_london}).

\begin{figure}[t!]
    \centering
    \begin{subfigure}{0.49\columnwidth}
        \centering
        \includegraphics[width=1\columnwidth]{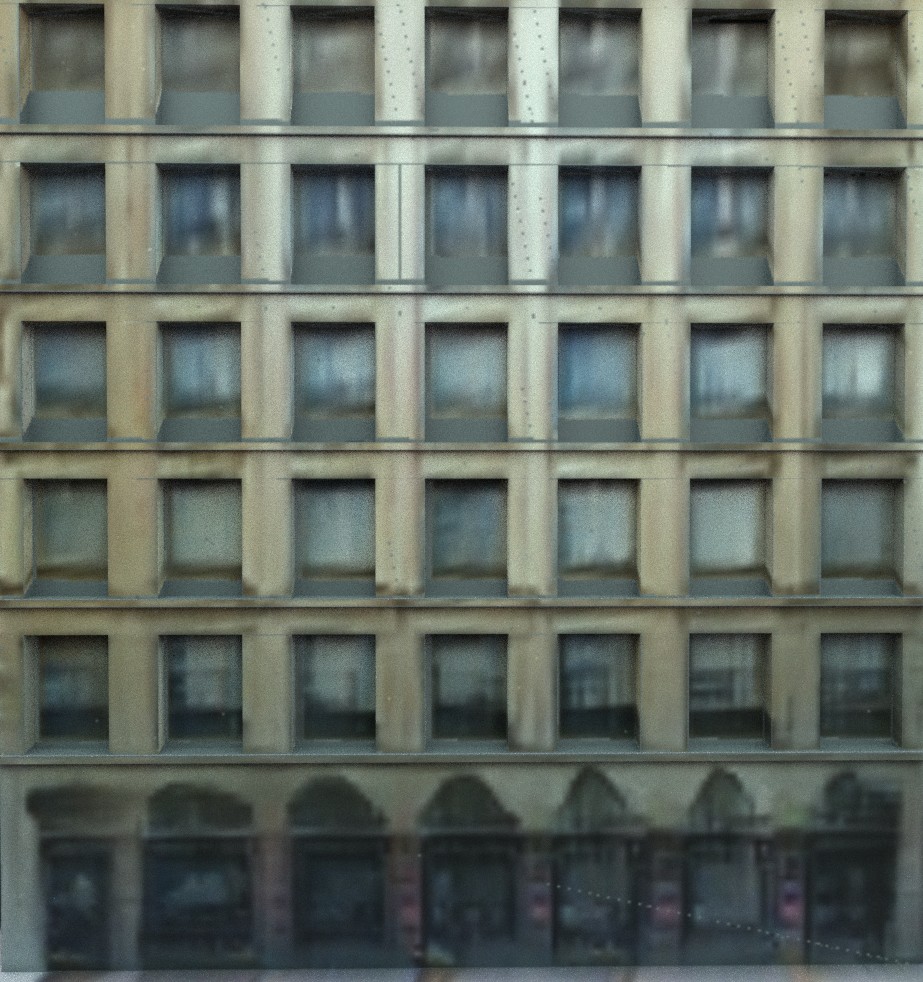}
    \end{subfigure}
    \begin{subfigure}{0.49\columnwidth}
        \centering
        \includegraphics[width=1\columnwidth]{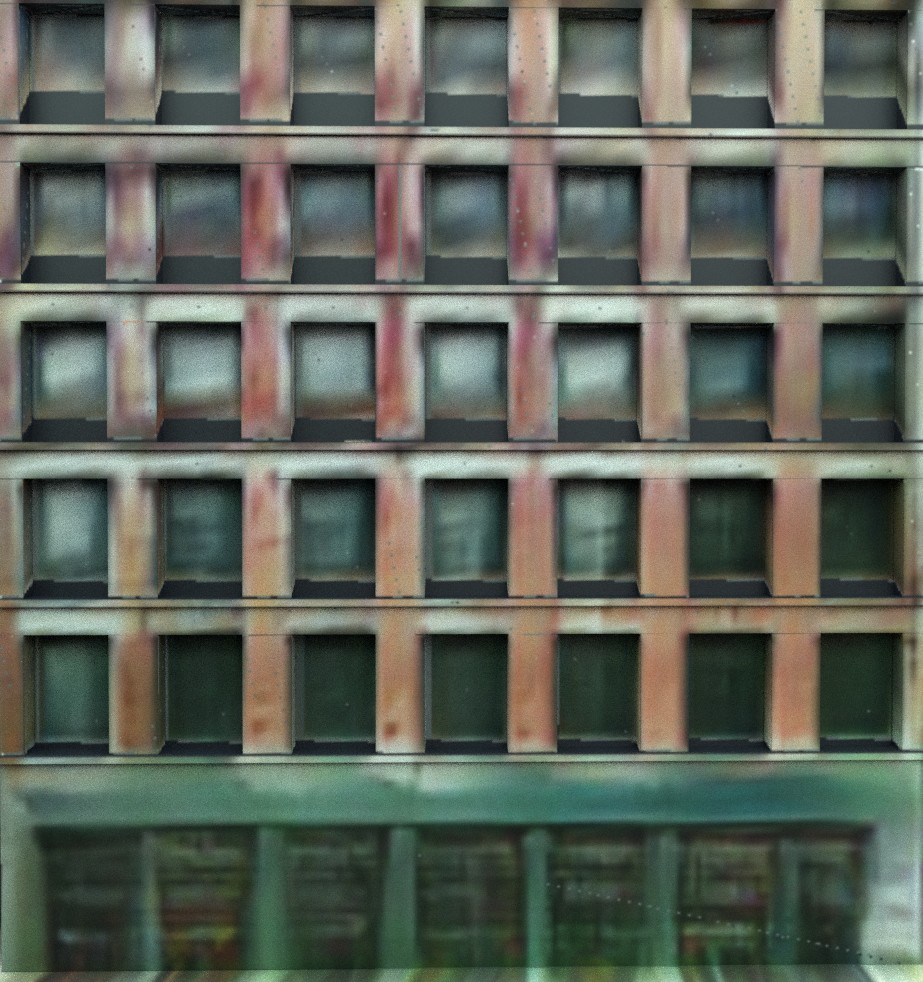}
    \end{subfigure}
    \hfill
    \begin{subfigure}{0.49\columnwidth}
        \centering
        \includegraphics[width=1\columnwidth]{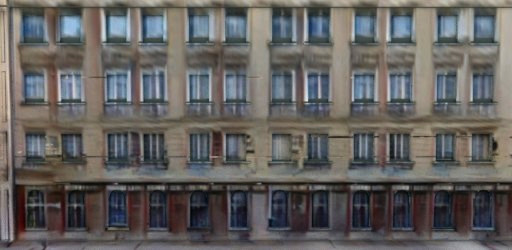}
    \end{subfigure}
    \begin{subfigure}{0.49\columnwidth}
        \centering
        \includegraphics[width=1\columnwidth]{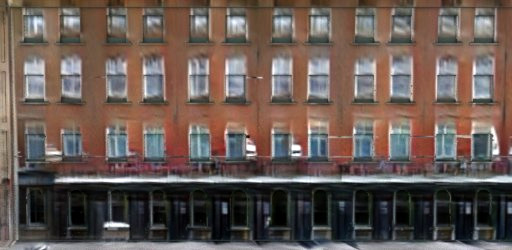}
    \end{subfigure}
    \hfill
    \begin{subfigure}{0.49\columnwidth}
        \centering
        \includegraphics[width=1\columnwidth]{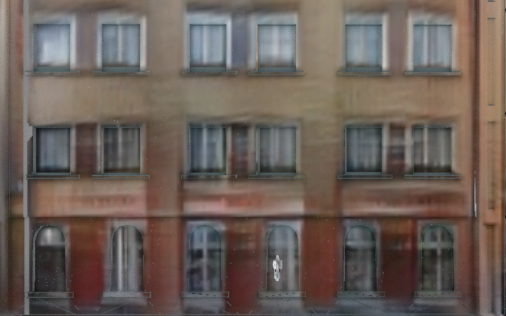}
    \end{subfigure}
    \begin{subfigure}{0.49\columnwidth}
        \centering
        \includegraphics[width=1\columnwidth]{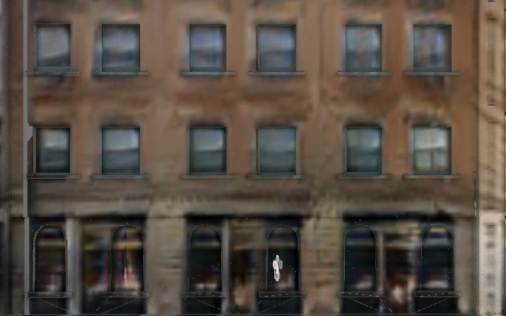}
    \end{subfigure}
    \hfill
    \vspace{-3mm}
    \caption{Street-level renderings of buildings in our test scene textured by our method trained on London images (left) and New York images (right). %Notice how textures generated by New York training data are more reddish compared to those of London training data which are more yellowish.
    }
    \label{fig:style_lo_ny}
    \vspace{-6mm}
\end{figure}

\vspace{-5mm}
\paragraph*{City style transfer.} 
% \melinos{need a better title here} 
Our method can be trained on street-level panoramic images collected from different cities in the world to give urban 3D models the appearance of a city's distinctive style, as illustrated by our London and New York examples in Figure \ref{fig:style_lo_ny}.
The generated textured city blocks appear distinctively textured with our pipeline trained on New York versus the ones textured with London panoramas. For example, a red brick-like appearance is visible in the walls of the meshes textured by our network trained in New York panoramas; such appearance is common in New York buildings. In contrast, meshes textured by our network trained on London panoramas appear with white, yellow and brown distributions of color on their facades, a color distribution which is common around London. 
%In Figure \ref{fig:3D_facade_comp} we observe the different styles between London and New-York, and verify the color and style of generated panoramas match different distributions based on the city that it was trained on. 
% We provide additional comparisons showing real-world panoramas and panoramic images rendered from our textured city blocks from London and New York in Figures \ref{fig:real_fake_comparison} and \ref{fig:real_fake_comparison_NY}. We can observe that the generated outputs follow color  distributions of given cities. Looking at the real-world panoramas, we observe the differences between the two cities: New York's facades often consist of gray-red shades of colors while London' facades tend to be brighter with yellow to blown color distributions. These distributions  are also encountered in our generated textures.

\begin{figure}[t!]
     \centering
     \begin{subfigure}[b]{0.49\textwidth}
        \centering
        \includegraphics[width=\textwidth]{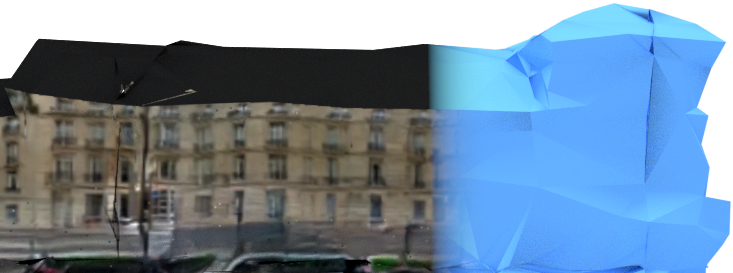}
     \end{subfigure}
     \hfill
     \begin{subfigure}[b]{0.49\textwidth}
        \centering
        \includegraphics[width=\textwidth]{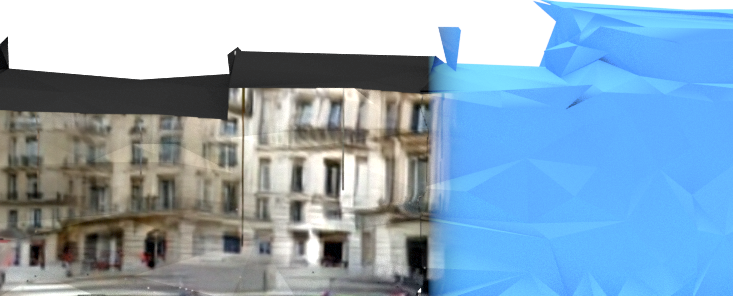}
    \end{subfigure}
    \vspace{-6mm}
    \caption{Our method can generate textures for challenging polygon meshes such as those from Google Earth. Here we show a street-level rendering of  a Google Earth mesh textured by our method trained on London panoramas. The mesh geometry is noisy with non-planar facades lacking window and door cues  (right). Still  our method manages to create an approximate mesh texture with such cues.}
    \label{fig:google_text}
    \vspace{-5mm}
\end{figure}

\vspace{-5mm}
\paragraph*{Mesh type generalization.}
We also demonstrate our pipeline on a  city 3D model from Google Earth, shown in Figure~\ref{fig:google_text}. 
%Note that the mesh does not contain any facade or window geometric information in contrast to the procedurally generated mesh. 
%\name creates reasonable quality textures, as illustrated in Figure \ref{fig:google_text}. 
Our method generates consistent facade textures, even for such  challenging polygon soups that are noisy and do not contain any facade or window geometric information, as shown in the untextured rendering of the same model (Figure~\ref{fig:google_text}, right).
%as illustrated in Figure \ref{fig:google_text}. %considering the style and the windows ordering. 
However, the absence of accurate facade geometry limits the diversity of texturing.

\begin{figure}[t!]
     \centering
    \includegraphics[width=0.5\textwidth, height=7cm]{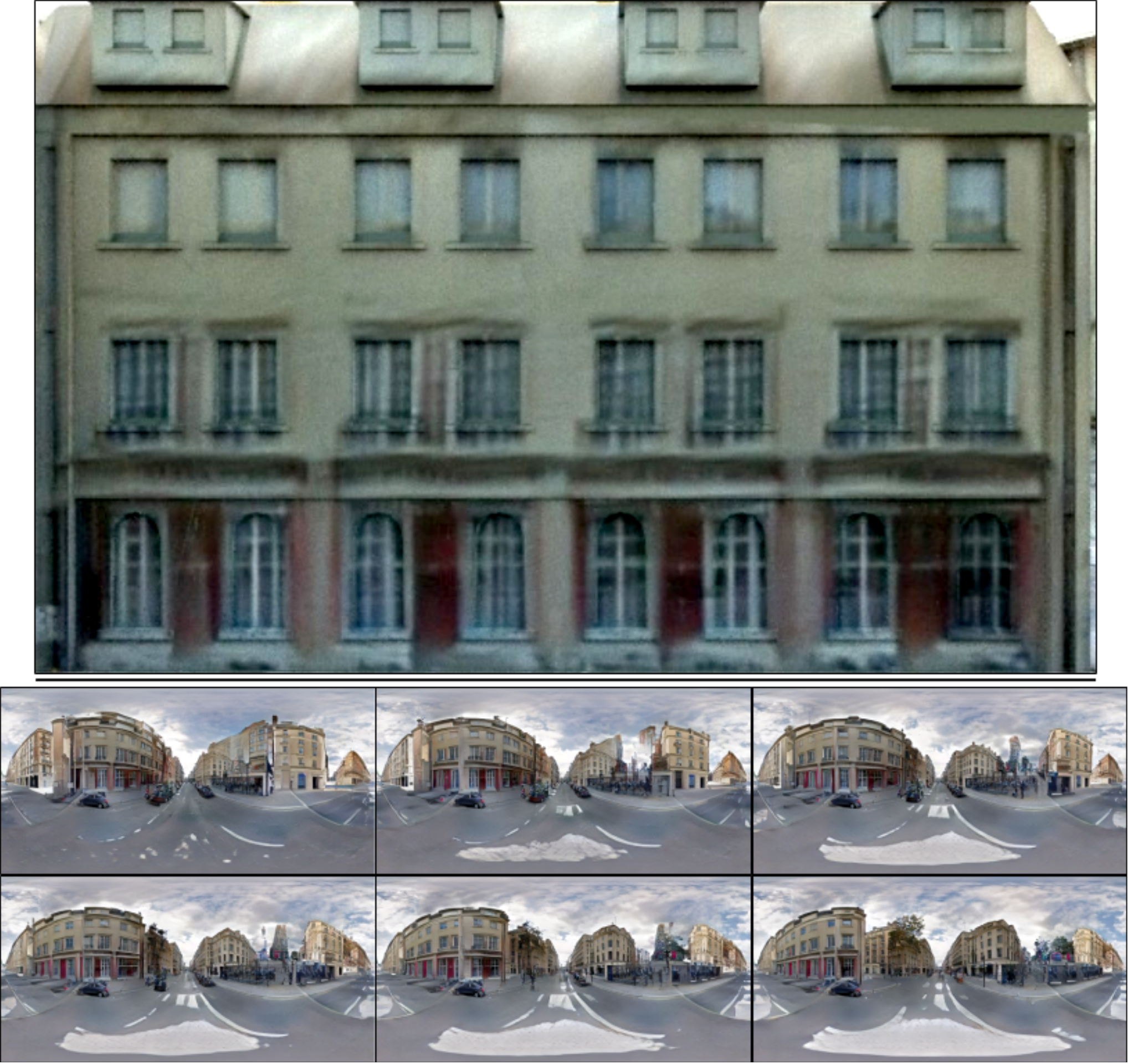}
        %  \caption{$y=x$}
        %  \label{fig:y equals x}
    \vspace{-6mm}
    \caption{A sequence of six consecutive panoramic outputs of our network trained on London data (bottom). Notice how each consecutive output tends to be more consistent with the previous, which allows our texture propagation module to better preserve some details and better align features such as windows and doors on facade geometry (top).}
    \label{fig:pano2facade}
    \vspace{-4mm}
\end{figure}

\vspace{-5mm}
\paragraph*{Inter-frame consistency.}
% \melinos{why just six}\yiangos{base on the facade length it needs the corresponding number of panos, we can show more or less panos it make no difference}
We qualitatively evaluate the degree of consistency between consecutive output textures from our architecture 
in Figure~\ref{fig:pano2facade}. We display six consecutive intermediate panoramic outputs of our image translation network trained on the London dataset. We note that each consecutive output tends to be consistent with the preceding images.
This helps to preserve some details in the generated textures, as seen at the top of Figure
\ref{fig:pano2facade}, which shows a close-up rendering of the textured 3D mesh on the right side of the street.
% Figure \ref{fig:consistency} also shows how partially textured images are translated into fully textured ones, helping with the  consistency of the generated texture. The overall appearance of the already textured part of the mesh smoothly blends with the untextured parts.
%\melinos{not sure about this image and what we are trying to show - maybe merge to \ref{fig:london_galery} ? and make that one larger by showing both final (output) and input panoramas?}. 

\begin{figure}[t!]
    \centering
    \includegraphics[width=1\columnwidth]{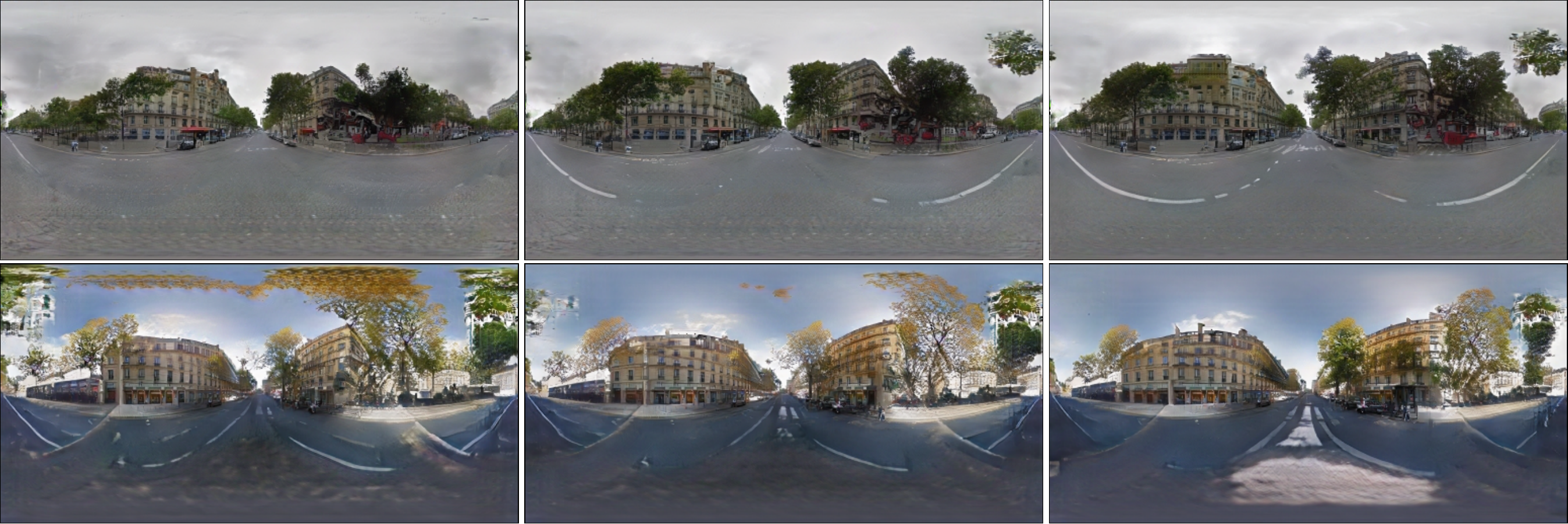}
    \vspace{-6mm}
    \caption{Tree artifacts appear on the facades in images generated from \textit{CUT} (top row). Our inter-frame consistency loss reduces these artifacts (bottom row).}
    \label{fig:tree_art}
    \vspace{-4mm}
\end{figure}

\begin{figure}[t!]
    \centering
    \includegraphics[width=1\columnwidth]{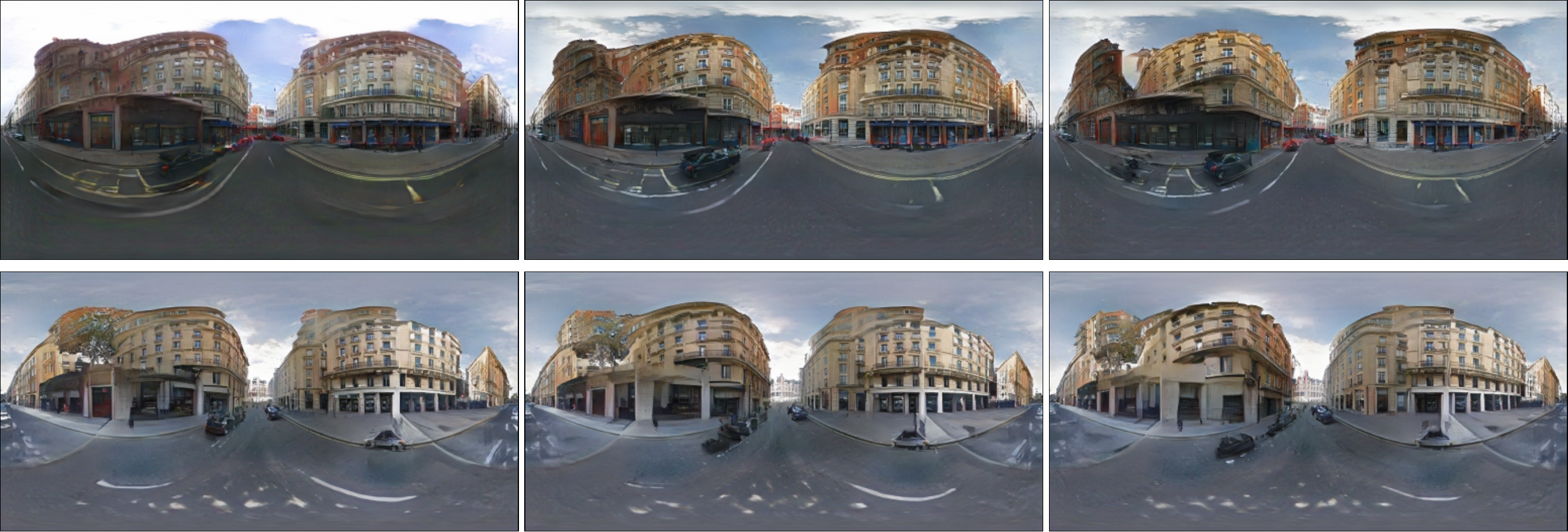}
    \vspace{-6mm}
    \caption{Style discontinuities occur between consecutive images generated from \textit{CUT} (top row) in contrast with \name which generates more consistent colors (bottom row).}
    \label{fig:Style_art}
    \vspace{-5mm}
\end{figure}

% \paragraph{Comparisons with the original CUT.} Figures \ref{fig:tree_art} and \ref{fig:Style_art} (top) shows results of the original CUT method: the method was trained on London data to learn the mapping between the domain of grayscale renderings and real-world panoramas directly i.e., without using partial textured images, and without the inter-frame consistency loss. We observe that
%  consecutive panoramas often differ due to tree artifacts suddenly appearing. In \ref{fig:tree_art} we observe that given the partially textured panorama the pipeline prevents generative undesirable tree artifacts on the following frames. The color of the facade in the third panorama in \ref{fig:Style_art} (top) differs from the first generated panorama (top). On the other hand, our network generates panoramas in a more consistent manner \ref{fig:tree_art} and \ref{fig:Style_art} (bottom) and by preventing additional noise.
% Figure \ref{fig:3D_facade_comp} shows differences of the original CycleGAN and our method in the output textured buildings, after applying our texture propagation step for both methods. Half of the textured building generated by the original CycleGAN has green-like appearance (e.g., grass- or tree-texture), which makes it inconsistent with the other half. In contrast, our textured building has a more consistent appearance. 
\vspace{-4mm}
% \paragraph{Comparison with \textit{CUT} and \textit{FrankenGAN}.}
\paragraph{Quantitative evaluation.}
\vspace{-1mm}
We first compare \name to the \textit{CUT} network when generating texture atlases. We note that \textit{CUT} does not condition its image translation network on the partially textured image and does not use our inter-frame consistency loss.
Both networks were trained on the London dataset. For fair comparison, we use the same equirectangular convolution for the first convolution layer of both architectures.
Table~\ref{tab:cut_vs_put} shows FID scores for panoramic renderings produced by  \name and \textit{CUT} for our test procedural scene. Since a large area of panoramic images consists of road and sky with little or no variation in texture, we also evaluate the FID on cropped versions of these renderings that isolate the facades from the roads and sky; we call this variant score as ``crop FID''. 
%As a baseline (\textit{orig. CUT}) we use the same model architecture trained in an unconditional manner. 
We additionally show the performance of \name and \emph{CUT} using three different blending approaches: \textit{no blend} where no blending is applied between frames, \textit{average} blending which takes the average RGB values between frames and our texture propagation (without weights), \textit{weighted} uses the weighted averaging scheme of Equation \ref{eq:interp}.
%approach as described in \ref{subsec:textprob}. 
%The importance of our inter-frame consistency loss and the conditional setting in large-scale mesh texturing illustrated in the Table~\ref{tab:cut_vs_put}. 
Our model outperforms \textit{CUT} regardless of the blending approach. 
%for a perceptual evaluation where \name received 111 persistent votes (56.9\%) in contrast to CUT which received 64 votes (32.8 \%) (additional details in supplementary).}}  
Note that, our texture propagation approach based on our weighted averaging scheme outperforms the no-blend or average-blend baselines.

We also compare our model to \textit{FrankenGAN} \cite{kelly2018frankengan} using the author-provided trained model. Note that blending cannot be used in FrankenGAN since it generates  textures for 3D\ objects individually (e.g., facade, window,  or roof). FrankenGAN has the additional benefit of semantic labels for\ these objects. In contrast, we do not assume that such labels are available for test meshes.  As shown in Table \ref{tab:cut_vs_put}, our approach still outperforms FrankenGAN.

\begin{table}[h!]
\centering
\begin{adjustbox}{max width=0.5\textwidth}
    \begin{tabular}{||@{}c@{}|@{}c@{}|@{}c@{}|@{}c@{}|@{}c@{}|@{}c@{}|@{}c@{}|} 
        \hline
         & \multicolumn{2}{| c |}{\textbf{No-Blend}$\downarrow$} &\multicolumn{2}{| c |}{\textbf{Average}$\downarrow$} & \multicolumn{2}{| c |}{\textbf{Weighted}$\downarrow$} \\ [0.5ex] 
        \hline
        \hline
        Method & full & crop & full & crop & full & crop \\ 
        \hline
        \textbf{\,CUT\,} & \,131.6\, & \,110.2\, & \,135.2\, & \,113.4\, & \,121.7\, & \,101.1\, \\
        \hline
        \textbf{\,FrankenGAN\,} & \,174.3\, & \,135.8\, & \,-\, & \,-\, & \,-\, & \,-\,\\
        \hline
        \textbf{\name} & \textbf{\,128.2\,} & \textbf{\,95.6\,} & \textbf{\,132.6\,} & \textbf{\,97.3\,} & \textbf{\,115.4\,} & \textbf{\,86.3\,} \\
        \hline          
    \end{tabular}
    \end{adjustbox}
    \vspace{-2mm}
    \caption{FID comparisons between \name and alternatives.}
    \label{tab:cut_vs_put}
\end{table}

%\vspace{-5mm}

In Figures \ref{fig:tree_art} and \ref{fig:Style_art} we show qualitative comparisons between \name and \textit{CUT}.
Figure~\ref{fig:tree_art} (top) shows failure cases of \textit{CUT}, i.e. undesired tree artifacts on the facades which \name{} decreases (bottom) by using the inter-frame consistency loss.
%the generation of undesirable tree artifacts on the facade area of the following viewpoints.
Moreover, the style of the facade between the first and the third panorama produced by \textit{CUT} differs in Figure \ref{fig:Style_art} (top) in contrast to \name which generates more consistent results (bottom). 

Finally, we performed a perceptual user study which showed that users preferred results generated from \name much more compared to the results from \textit{CUT}; see supplementary materials for additional details.

%On the other hand, our network generates panoramas in consistent manner as illustrated at Figures \ref{fig:tree_art} and \ref{fig:Style_art} (bottom row).
% Figures \ref{fig:tree_art} and \ref{fig:Style_art} (top) shows results of the original CUT method: the method was trained on London data to learn the mapping between the domain of grayscale renderings and real-world panoramas directly i.e., without using partial textured images, and without the inter-frame consistency loss. We observe that
%  consecutive panoramas often differ due to tree artifacts suddenly appearing. In \ref{fig:tree_art} we observe that given the partially textured panorama the pipeline prevents generative undesirable tree artifacts on the following frames. The color of the facade in the third panorama in \ref{fig:Style_art} (top) differs from the first generated panorama (top). On the other hand, our network generates panoramas in a more consistent manner \ref{fig:tree_art} and \ref{fig:Style_art} (bottom) and by preventing additional noise.
\vspace{-5mm}
\paragraph{Ablation study.}
We also designed an ablation study to support the three main design choices for our network architecture, namely the use of the mask $M$ in the inter-frame consistency loss (Equation \ref{eq:cons}),
%The area of consistency image (\textit{Part consistency}) that was used to compute inter-frame loss, 
the use of equirectangular convolution for the first convolutional layer of the network~\cite{equi_conv}, and the merging of the untextured (grayscale) and partially textured image to create $3$-channel inputs.
%and the comparison between our conditional (\name) (i.e., conditioned on previously partially textured inputs),
%and unconditional texturing approach (\textit{Orig. CUT}). 
%In our evaluation we use the FID score\cite{fid_paper}, which is a measure of distance between feature vectors extracted from real and generated images. 
%For each variation of our network trained on the London dataset, we compute the FID score on a set of street-level renderings (panoramic, with global illumination) of our test procedurally-generated scene, as textured by our method.
%Low scores indicate that feature statistics in the generated images agree more with ones in real images. 
%A textured skybox is used on outputs to avoid empty background in generated images that could affect the FID score. 
%To get fair and targeted metrics regarding mesh texturing quality, and avoid undesirable aspects(such as generated sky) to alter the evaluation we re-render panoramic images given the generated textures. For the re-rendering a regular blender's sky texture is used. 
%Additionally, since a large area of panoramic images consists of road and sky with little or no variation in texture, we evaluate the FID on cropped versions of these renderings, that isolate the facades from the roads and sky.
%focus on facades texture quality we compared our models given 
%We compare the FID score between conditional and unconditional texture generation, different blending approaches and \name variations.

%\paragraph{Ablation Study.}
Table~\ref{tab:ablation_study} reports the ``crop FID'' scores for five variations of \name produced by combination of the three design choices we made. 
Each row is a different model, incorporating different variations of the three choices.
Comparing \nameA and \nameB we can see that our choice to merge untextured and partially textured images in a 3-channel input improves FID scores.
Adding equirectangular convolution in the first convolution layer of \nameC slightly improves FID score, while incorporating the mask in the inter-frame consistency in \nameD gives a much larger improvement compared to \nameB. Finally, using all three modifications results in the best FID score for the full model.
%by merging the two images we reduce memory requirements and improves the image quality (i.e. improved FID score). Equirectangular convolutions seem to better capture image information for panoramic images in contrast to regular convolution as can be seen, \nameC improves by $0.232$ compared to \nameB and \nameE by $1.262$ compared to \nameD. Moreover, by changing the strategy of how we compute inter-frame consistency loss (from whole image to only the visible part of the input image) led to considerable improvement between \nameB - \nameD and \nameC - \nameE by $4.903$ and $5.93$ respectively. We refer to \nameE as \name for the rest of the paper.
 
\begin{table}[th!]
\begin{adjustbox}{max width=0.5\textwidth}
%   \begin{center}
    % 
    \begin{tabular}{||@{}c@{}|@{}c@{}|@{}c@{}|@{}c@{}|@{}c@{}|@{}c@{}||}
    \hline
      \textbf{Model} & \textbf{\,Masked Cons.\,} & \textbf{\,Equir. Conv.\,} & \textbf{\,Gray+RGB\,} & \textbf{\,crop FID\,}$\downarrow$\\
      \hline\hline
      \nameA  & \textendash & \textendash & \textendash & 93.10 \\
      \nameB  & \textendash & \textendash & \checkmark  & 92.47 \\
      \nameC  & \textendash & \checkmark & \checkmark & 
      92.23 \\
      \nameD  & \checkmark & \textendash & \checkmark & 
      87.56 \\
      \,full \name\,  & \checkmark & \checkmark & \checkmark & \textbf{86.30} \\
      \hline
    \end{tabular}
%   \end{center}https://www.overleaf.com/project/6044b29047582896400943bb
\end{adjustbox}
\vspace{-2mm}
\caption{FID scores for \name design choices. ``Masked Cons.'' means whether we use the mask in the consistency loss, ``Equir. Conv.'' means whether we use equirectangular convolution, ``Gray+RGB'' means whether we merge the grayscale rendering with the RGB partially textured image in a 3-channel image (or treat it as 4-channel image).}
\label{tab:ablation_study}
\vspace{-4mm}
\end{table}

\section{Limitations and Conclusion}
We  presented \name, a method for texturing urban 3D scenes.
Our method is able to texture large  urban meshes by training  on unpaired panoramic image  and 3D\ geometry.
One limitation of \name is its inability to texture roofs;  incorporating aerial images into our pipeline can help. 
Another limitation  affecting the output texture resolution is that
%, compared to city texturing approaches which tile and repeat the same texture over hundreds of locations, 
storing our large atlas is memory intensive. 
%Recent advances in rendering pipelines have made it possible handle such large textures, %while consumer-level GPUs offer ever increasing amounts of dedicated memory. 
\name can also be challenged by  large domain gaps in the distributions of features between panoramic photos and geometry. 
%These lead to texture artifacts, such as red areas over the streets and green shading over the facades. While we found CUT's contrastive consistency loss flexible to smaller disparities in such distributions, but if the domain gap was sufficiently large, such artifacts became evident. 
A related issue is the reconstruction of artifacts due to   lens aberrations and shadows in panoramas. 
%While some of these features, such as shadows, can be useful, the lack of control is problematic.
In the future, we would like to allow users to explore the style-space of an urban environment  by navigating a latent space.  Another direction is to increase our texture resolution to better portray fine details using super-resolution neural architectures. %In particular, we would like to vary the resolution based on the viewpoint, to generate millimeter resolution textures close to the camera, and consistent cityscapes in the far distance, all in real-time. 
\vspace{-5mm}
\paragraph{Acknowledgements.} Our work was funded by the ERDF and the Republic of Cyprus through
the RIF (Project EXCELLENCE/1216/0352),  the EU
H2020 Research and Innovation Programme, and the Cyprus Deputy Ministry of Research,
Innovation and Digital Policy (Grant  739578).

%%%%%%%%% 
{\small
\bibliographystyle{ieee_fullname}
\bibliography{egbib}

\begin{thebibliography}{10}\itemsep=-1pt

\bibitem{anguelov2010google}
Dragomir Anguelov, Carole Dulong, Daniel Filip, Christian Frueh, St{\'e}phane
  Lafon, Richard Lyon, Abhijit Ogale, Luc Vincent, and Josh Weaver.
\newblock Google street view: Capturing the world at street level.
\newblock {\em Computer}, 43(6):32--38, 2010.

\bibitem{bohm2004multi}
Jan B{\"o}hm.
\newblock Multi-image fusion for occlusion-free fa{\c{c}}ade texturing.
\newblock {\em International Archives of the Photogrammetry, Remote Sensing and
  Spatial Information Sciences}, 35(5):867--872, 2004.

\bibitem{bornik2001high}
Alexander Bornik, Konrad Karner, Joachim Bauer, Franz Leberl, and Heinz Mayer.
\newblock High-quality texture reconstruction from multiple views.
\newblock {\em The journal of visualization and computer animation},
  12(5):263--276, 2001.

\bibitem{chang2003synthesis}
Yao-Xun Chang and Zen-Chung Shih.
\newblock The synthesis of rust in seawater.
\newblock {\em The Visual Computer}, 19(1):50--66, 2003.

\bibitem{chen2005visual}
Yanyun Chen, Lin Xia, Tien-Tsin Wong, Xin Tong, Hujun Bao, Baining Guo, and
  Heung-Yeung Shum.
\newblock Visual simulation of weathering by $\gamma$-ton tracing.
\newblock In {\em {ACM SIGGRAPH}}, pages 1127--1133. 2005.

\bibitem{choi2018stargan}
Yunjey Choi, Minje Choi, Munyoung Kim, Jung-Woo Ha, Sunghun Kim, and Jaegul
  Choo.
\newblock Stargan: Unified generative adversarial networks for multi-domain
  image-to-image translation.
\newblock In {\em Proceedings of the IEEE conference on computer vision and
  pattern recognition}, pages 8789--8797, 2018.

\bibitem{blender}
Blender~Online Community.
\newblock {\em Blender - a 3D modelling and rendering package}.
\newblock Blender Foundation, Stichting Blender Foundation, Amsterdam, 2018.

\bibitem{dai2013example}
Dengxin Dai, Hayko Riemenschneider, Gerhard Schmitt, and Luc Van~Gool.
\newblock Example-based facade texture synthesis.
\newblock pages 1065--1072, 2013.

\bibitem{darabi2012image}
Soheil Darabi, Eli Shechtman, Connelly Barnes, Dan~B Goldman, and Pradeep Sen.
\newblock Image melding: combining inconsistent images using patch-based
  synthesis.
\newblock {\em {ACM SIGGRAPH}}, 31(4):1--10, 2012.

\bibitem{efros2001image}
Alexei~A Efros and William~T Freeman.
\newblock Image quilting for texture synthesis and transfer.
\newblock In {\em Proceedings of the 28th annual conference on Computer
  graphics and interactive techniques}, pages 341--346, 2001.

\bibitem{efros1999texture}
Alexei~A Efros and Thomas~K Leung.
\newblock Texture synthesis by non-parametric sampling.
\newblock In {\em {IEEE ICCV}}, volume~2, pages 1033--1038. IEEE, 1999.

\bibitem{Esri:2020:CE}
Esri.
\newblock Esri {CityEngine} 2020.1, 2020.

\bibitem{FanCompletion}
Lubin Fan, Przemyslaw Musialski, Ligang Liu, and Peter Wonka.
\newblock Structure completion for facade layouts.
\newblock {\em {ACM SIGGRAPH Asia}}, 33(6), Nov. 2014.

\bibitem{equi_conv}
Clara Fernandez-Labrador, Jose~M. Facil, Alejandro Perez-Yus, Cendric
  Demonceaux, Javier Civera, and Jose~J. Guerrero.
\newblock Corners for layout: End-to-end layout recovery from 360 images.
\newblock 2019.

\bibitem{fruhstuck2019tilegan}
Anna Fr{\"u}hst{\"u}ck, Ibraheem Alhashim, and Peter Wonka.
\newblock Tilegan: synthesis of large-scale non-homogeneous textures.
\newblock {\em {ACM SIGGRAPH}}, 38(4):1--11, 2019.

\bibitem{gatys2015texture}
Leon Gatys, Alexander~S Ecker, and Matthias Bethge.
\newblock Texture synthesis using convolutional neural networks.
\newblock In {\em {NIPS}}, pages 262--270, 2015.

\bibitem{gatys2016image}
Leon~A Gatys, Alexander~S Ecker, and Matthias Bethge.
\newblock Image style transfer using convolutional neural networks.
\newblock In {\em Proceedings of the IEEE conference on computer vision and
  pattern recognition}, pages 2414--2423, 2016.

\bibitem{guerin2017interactive}
{\'E}ric Gu{\'e}rin, Julie Digne, {\'E}ric Galin, Adrien Peytavie, Christian
  Wolf, Bedrich Benes, and Beno{\^\i}t Martinez.
\newblock Interactive example-based terrain authoring with conditional
  generative adversarial networks.
\newblock {\em {ACM TOG}}, 36(6):1--13, 2017.

\bibitem{guo2019gan}
Xi Guo, Zhicheng Wang, Qin Yang, Weifeng Lv, Xianglong Liu, Qiong Wu, and Jian
  Huang.
\newblock Gan-based virtual-to-real image translation for urban scene semantic
  segmentation.
\newblock {\em Neurocomputing}, 2019.

\bibitem{PE:VMV:VMV12:063-070}
Tobias Günther, Kai Rohmer, and Thorsten Grosch.
\newblock {GPU-accelerated Interactive Material Aging}.
\newblock In Michael Goesele, Thorsten Grosch, Holger Theisel, Klaus Toennies,
  and Bernhard Preim, editors, {\em Vision, Modeling and Visualization}. The
  Eurographics Association, 2012.

\bibitem{hedman2018deep}
Peter Hedman, Julien Philip, True Price, Jan-Michael Frahm, George Drettakis,
  and Gabriel Brostow.
\newblock Deep blending for free-viewpoint image-based rendering.
\newblock {\em {ACM TOG}}, 37(6):1--15, 2018.

\bibitem{henderson2020leveraging}
Paul Henderson, Vagia Tsiminaki, and Christoph~H Lampert.
\newblock Leveraging 2d data to learn textured 3d mesh generation.
\newblock pages 7498--7507, 2020.

\bibitem{henzler2020neuraltexture}
Philipp Henzler, Niloy~J Mitra, , and Tobias Ritschel.
\newblock Learning a neural 3d texture space from 2d exemplars.
\newblock In {\em {IEEE CVPR}}, June 2019.

\bibitem{huang2020adversarial}
Jingwei Huang, Justus Thies, Angela Dai, Abhijit Kundu, Chiyu~Max Jiang,
  Leonidas Guibas, Matthias Nie{\ss}ner, and Thomas Funkhouser.
\newblock Adversarial texture optimization from rgb-d scans.
\newblock {\em arXiv preprint arXiv:2003.08400}, 2020.

\bibitem{huang2018multimodal}
Xun Huang, Ming-Yu Liu, Serge Belongie, and Jan Kautz.
\newblock Multimodal unsupervised image-to-image translation.
\newblock In {\em Proceedings of the European Conference on Computer Vision
  (ECCV)}, pages 172--189, 2018.

\bibitem{pix2pix}
Phillip Isola, Jun-Yan Zhu, Tinghui Zhou, and Alexei~A Efros.
\newblock Image-to-image translation with conditional adversarial networks.
\newblock {\em {IEEE CVPR}}, 2017.

\bibitem{Johnson}
Justin Johnson, Alexandre Alahi, and Fei fei Li.
\newblock Perceptual losses for real-time style transfer and super-resolution.
\newblock In {\em {ECCV}}, 2016.

\bibitem{kato2018neural}
Hiroharu Kato, Yoshitaka Ushiku, and Tatsuya Harada.
\newblock Neural 3d mesh renderer.
\newblock In {\em Proceedings of the IEEE Conference on Computer Vision and
  Pattern Recognition}, pages 3907--3916, 2018.

\bibitem{kelly2018frankengan}
Tom Kelly, Paul Guerrero, Anthony Steed, Peter Wonka, and Niloy~J Mitra.
\newblock Frankengan: guided detail synthesis for building mass models using
  style-synchonized gans.
\newblock {\em {ACM TOG}}, 37(6):1--14, 2018.

\bibitem{kelly2011interactive}
Tom Kelly and Peter Wonka.
\newblock Interactive architectural modeling with procedural extrusions.
\newblock {\em {ACM TOG}}, 30(2):14, 2011.

\bibitem{klingner2013street}
Bryan Klingner, David Martin, and James Roseborough.
\newblock Street view motion-from-structure-from-motion.
\newblock In {\em Proceedings of the IEEE International Conference on Computer
  Vision}, pages 953--960, 2013.

\bibitem{li2015approximate}
Chuan Li and Michael Wand.
\newblock Approximate translational building blocks for image decomposition and
  synthesis.
\newblock {\em ACM Transactions on Graphics (TOG)}, 34(5):1--16, 2015.

\bibitem{LiangPatch}
Lin Liang, Ce Liu, Ying-Qing Xu, Baining Guo, and Heung-Yeung Shum.
\newblock Real-time texture synthesis by patch-based sampling.
\newblock {\em ACM Trans. Graph.}, 20(3):127–150, July 2001.

\bibitem{lu2007context}
Jianye Lu, Athinodoros~S Georghiades, Andreas Glaser, Hongzhi Wu, Li-Yi Wei,
  Baining Guo, Julie Dorsey, and Holly Rushmeier.
\newblock Context-aware textures.
\newblock {\em ACM Transactions on Graphics (TOG)}, 26(1):3--es, 2007.

\bibitem{meshry2019neural}
Moustafa Meshry, Dan~B Goldman, Sameh Khamis, Hugues Hoppe, Rohit Pandey, Noah
  Snavely, and Ricardo Martin-Brualla.
\newblock Neural rerendering in the wild.
\newblock In {\em {IEEE CVPR}}, pages 6878--6887, 2019.

\bibitem{mirowski2019streetlearn}
Piotr Mirowski, Andras Banki-Horvath, Keith Anderson, Denis Teplyashin,
  Karl~Moritz Hermann, Mateusz Malinowski, Matthew~Koichi Grimes, Karen
  Simonyan, Koray Kavukcuoglu, Andrew Zisserman, et~al.
\newblock The streetlearn environment and dataset.
\newblock {\em arXiv preprint arXiv:1903.01292}, 2019.

\bibitem{Mueller:2006:PMB}
Pascal Mueller, Peter Wonka, Simon Haegler, Andreas Ulmer, and Luc Van~Gool.
\newblock Procedural modeling of buildings.
\newblock {\em {ACM TOG}}, 25(3):614--623, 2006.

\bibitem{muller2007image}
Pascal M{\"u}ller, Gang Zeng, Peter Wonka, and Luc Van~Gool.
\newblock Image-based procedural modeling of facades.
\newblock {\em {ACM SIGGRAPH}}, 26(3):85, 2007.

\bibitem{niem1995mapping}
Wolfgang Niem and Hellward Broszio.
\newblock Mapping texture from multiple camera views onto 3d-object models for
  computer animation.
\newblock In {\em Proceedings of the International Workshop on Stereoscopic and
  Three Dimensional Imaging}, pages 99--105, 1995.

\bibitem{oechsle2019texture}
Michael Oechsle, Lars Mescheder, Michael Niemeyer, Thilo Strauss, and Andreas
  Geiger.
\newblock Texture fields: Learning texture representations in function space.
\newblock In {\em {IEEE ICCV}}, pages 4531--4540, 2019.

\bibitem{Parish:2001:PMC}
Yoav I.~H. Parish and Pascal M{\"u}ller.
\newblock Procedural modeling of cities.
\newblock In {\em Proceedings of SIGGRAPH 2001}, pages 301--308, 2001.

\bibitem{CUT}
Taesung Park, Richarf~Zhang Alexei A.~Efros, and Jun-Yan Zhu.
\newblock Contrastive learning for unpaired image-to-image translation.
\newblock {\em {ECCV}}, 2020.

\bibitem{Raj_2019_CVPR_Workshops}
Amit Raj, Cusuh Ham, Connelly Barnes, Vladimir Kim, Jingwan Lu, and James Hays.
\newblock Learning to generate textures on 3d meshes.
\newblock In {\em IEEE CVPR Workshops}, June 2019.

\bibitem{ronneberger2015u}
Olaf Ronneberger, Philipp Fischer, and Thomas Brox.
\newblock U-net: Convolutional networks for biomedical image segmentation.
\newblock In {\em International Conference on Medical image computing and
  computer-assisted intervention}, pages 234--241. Springer, 2015.

\bibitem{Schwarz:2015:APM}
Michael Schwarz and Pascal M{\"u}ller.
\newblock Advanced procedural modeling of architecture.
\newblock {\em {ACM TOG}}, 34(4):107:1--107:12, 2015.

\bibitem{sharma2019unsupervised}
Alisha Sharma and Jonathan Ventura.
\newblock Unsupervised learning of depth and ego-motion from cylindrical
  panoramic video.
\newblock {\em arXiv preprint arXiv:1901.00979}, 2019.

\bibitem{song2018im2pano3d}
Shuran Song, Andy Zeng, Angel~X Chang, Manolis Savva, Silvio Savarese, and
  Thomas Funkhouser.
\newblock Im2pano3d: Extrapolating 360 structure and semantics beyond the field
  of view.
\newblock In {\em {IEEE CVPR}}, pages 3847--3856, 2018.

\bibitem{wang2018video}
Ting-Chun Wang, Ming-Yu Liu, Jun-Yan Zhu, Guilin Liu, Andrew Tao, Jan Kautz,
  and Bryan Catanzaro.
\newblock Video-to-video synthesis.
\newblock {\em arXiv preprint arXiv:1808.06601}, 2018.

\bibitem{WeiTree}
Li-Yi Wei and Marc Levoy.
\newblock Fast texture synthesis using tree-structured vector quantization.
\newblock In {\em Proceedings of the 27th Annual Conference on Computer
  Graphics and Interactive Techniques}, SIGGRAPH ’00, page 479–488, USA,
  2000. ACM Press/Addison-Wesley Publishing Co.

\bibitem{wen2019pixel2mesh++}
Chao Wen, Yinda Zhang, Zhuwen Li, and Yanwei Fu.
\newblock Pixel2mesh++: Multi-view 3d mesh generation via deformation.
\newblock In {\em Proceedings of the IEEE International Conference on Computer
  Vision}, pages 1042--1051, 2019.

\bibitem{wu2016learning}
Jiajun Wu, Chengkai Zhang, Tianfan Xue, Bill Freeman, and Josh Tenenbaum.
\newblock Learning a probabilistic latent space of object shapes via 3d
  generative-adversarial modeling.
\newblock In {\em {NIPS}}, pages 82--90, 2016.

\bibitem{zhang2020synthesis}
X Zhang, C May, and D Aliaga.
\newblock Synthesis and completion of facades from satellite imagery.
\newblock 2020.

\bibitem{ZhouTextureSynthesis}
Yang Zhou, Zhen Zhu, Xiang Bai, Dani Lischinski, Daniel Cohen-Or, and Hui
  Huang.
\newblock Non-stationary texture synthesis by adversarial expansion.
\newblock {\em {ACM SIGGRAPH}}, 37(4), July 2018.

\bibitem{cycleGAN}
Zhu, Jun-Yan, Park Taesung, Isola, Phillip, and Efros~Alexei A.
\newblock Unpaired image-to-image translation using cycle-consistent
  adversarial networks.
\newblock {\em {IEEE ICCV}}, 2017.

\end{thebibliography}
}

\end{document}